\documentclass[runningheads]{llncs}

\usepackage{eccv}

\usepackage{eccvabbrv}

\usepackage{graphicx}
\usepackage{booktabs}

\usepackage[accsupp]{axessibility}  

\usepackage{multirow}

\usepackage{hyperref}

\usepackage{orcidlink}

\begin{document}

\title{LEGO: Leveled Language Gaussian Splatting} 
\titlerunning{LEGO: Leveled Language Gaussian Splatting}

\author{
Yuning Peng\inst{1}\orcidlink{0009-0007-5459-4691} \and
Haiping Wang\inst{1,2,\dagger}\orcidlink{0000-0002-8370-4585} \and
Yuan Liu\inst{2} \and
Yipeng Lu\inst{1} \and
Zhen Dong\inst{1}\orcidlink{0000-0002-0152-3300} \and
Bisheng Yang\inst{1}\orcidlink{0000-0001-7736-0803}
}

\authorrunning{Y.~Peng et al.}

\institute{Wuhan University\\
\email{\{yuningpeng,hpwang,luyipeng,dongzhenwhu,bshyang\}@whu.edu.cn}\\
\and
Hong Kong University of Science and Technology\\
\email{yuanly@connect.hku.hk}}

\maketitle
\renewcommand{\thefootnote}{†}
\footnotetext{Corresponding author.}

\begin{abstract}

We introduce LEGO for advanced open-vocabulary scene understanding.
Beyond basic concept recognition, its core innovation lies in capturing the intrinsic semantic hierarchies within the scene, such as the "flowerpot $\to$ bouquet $\to$ bud $\to$ petal" lineage.
While foundation models like SAM can identify multi-granular structures in 2D, their partitions are strictly perspective-bound and lack cross-view consensus.
LEGO self-adaptively re-grades volatile multi-view SAM granularities into a unified, 3D-consistent hierarchy. This provides precise supervision for the structurally coherent, multi-level segmentation of 3D scenes.
By grounding these segments with CLIP embeddings, LEGO recovers open-vocabulary semantic logic across hierarchical levels.
Furthermore, by incorporating spatial relationships, we elevate these segments into level-wise language scene graphs, effectively empowering Large Language Models to perform complex, context-aware spatial reasoning and precise visual grounding.
Experimental results demonstrate that LEGO establishes new state-of-the-art performance across both promptable and open-vocabulary 3D segmentation benchmarks, exhibiting advanced hierarchical scene decomposition and context-aware spatial reasoning. Project page: \url{https://pz0826.github.io/LEGO-Webpage/}
\end{abstract}

\begin{figure}[tb]
  \centering
  \includegraphics[width=\linewidth]{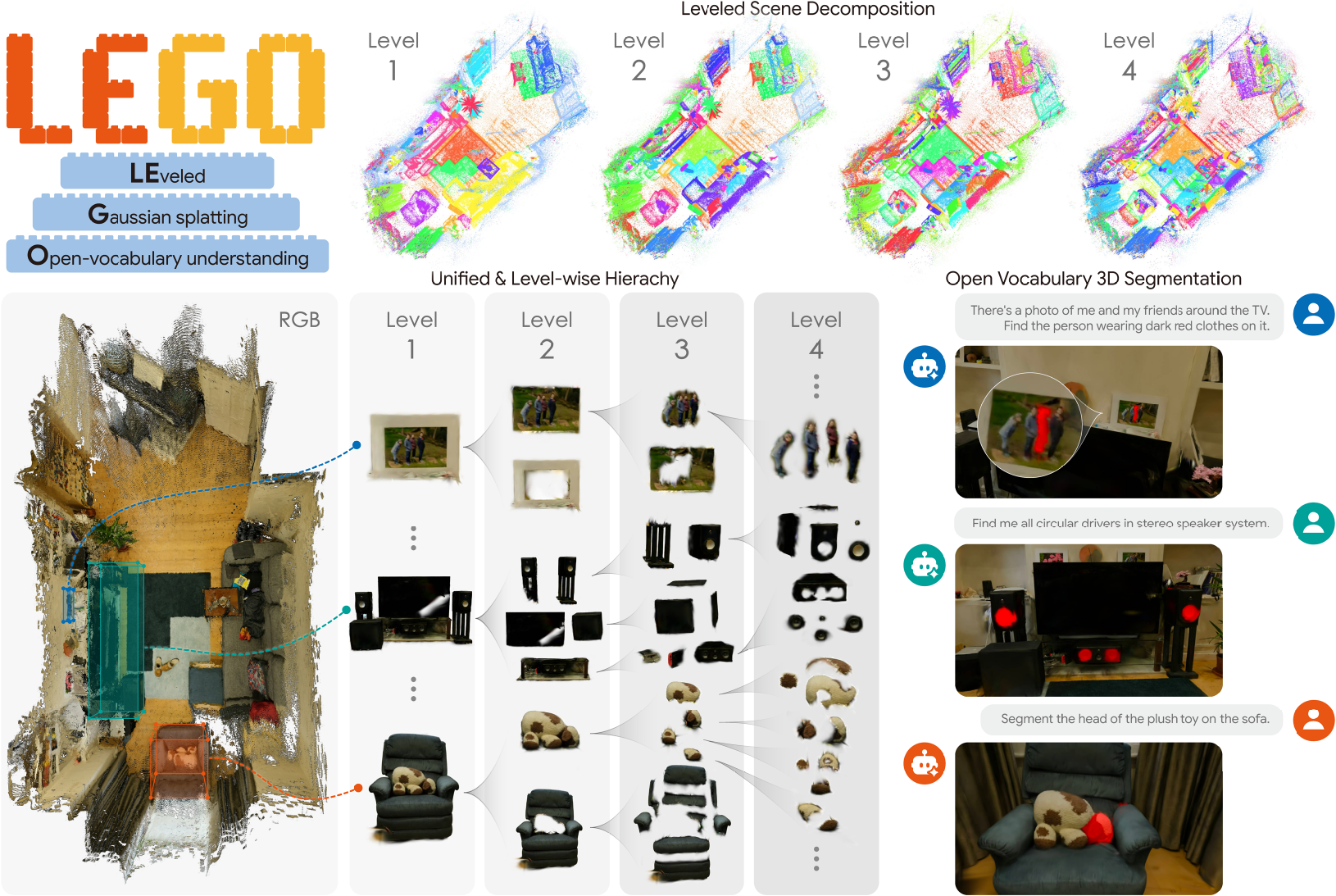}
  \caption{\textbf{LEveled Gaussian splatting for Open-vocabulary scene understanding (LEGO).} By parsing scenes into a unified, level-wise semantic hierarchy, LEGO establishes a robust foundation for multi-granular open-vocabulary understanding, seamlessly enabling automatic decomposition and context-aware spatial reasoning.
  }
  \label{fig:teaser}
\end{figure}

\begin{figure}[tb]
  \centering
  \includegraphics[width=\linewidth]{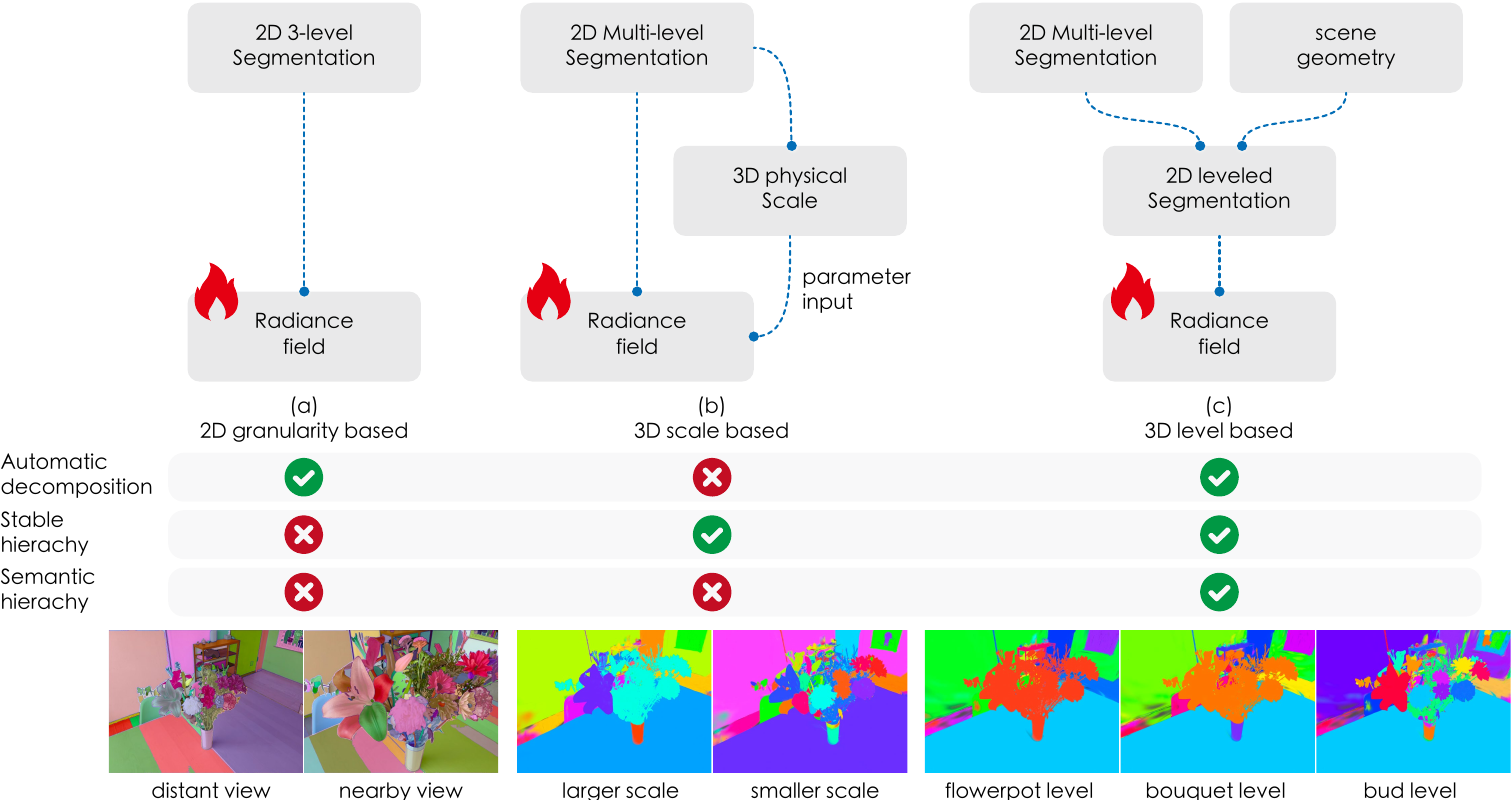}
  \caption{\textbf{Comparison of 3D hierarchical segmentation paradigms.} \textbf{(a) 2D Granularity-based} methods lift view-dependent masks, leading to inconsistent 3D hierarchies across different camera distances (\eg, petals in a nearby view vs. a whole bud in a distant view). \textbf{(b) 3D Scale-based} methods rely on absolute physical sizes, suffering from a semantic-scale gap. Due to intra-class scale variance, a single scale parameter may over-segment a large flower into petals while keeping smaller ones intact. \textbf{(c) Our 3D Level-based paradigm} automatically discovers a stable and consistent semantic hierarchy, grouping entities by their true structural compositional rank (\eg, flowerpot $\to$ bouquet $\to$ bud) regardless of viewing distance or absolute physical size.
  }
  \label{fig:intro_compare}
\end{figure}

\section{Introduction}
\label{sec:intro}

As 3D scene segmentation evolves from closed-set taxonomies to open-vocabulary understanding, a critical aspect is that \textit{real-world semantics are inherently hierarchical}. When humans perceive a scene, objects can be interpreted at various levels based on their semantic context, such as ``flowerpot $\to$ bouquet $\to$ bud $\to$ petal''. Therefore, a robust 3D open-vocabulary system must not only recognize arbitrary concepts but also explicitly construct the hierarchical semantic structures underlying the 3D space.

Segment Anything Model (SAM) \cite{kirillov2023sam} enables hierarchical 2D segmentation through three granularities: whole, part, and subpart. While lifting these masks as 3D levels is a seemingly direct path to building 3D hierarchies, existing methods face two fundamental limitations.

\textbf{View-Dependency in Granularity-based Distillation.}
The first category of work \cite{qin2024langsplat,zhan2025hilsplat,cheng2024occamlgs,dai2025thgs,cen2025laga} directly maps SAM’s 2D granularities to 3D hierarchies. However, SAM labels are inherently view-dependent, determined by an object’s relative granularity in the 2D image plane rather than its intrinsic 3D structure. For example, as shown in~\cref{fig:intro_compare}(a), a flower may be segmented into individual petals in a close-up but a complete bud from a distance at the same granularity. Distilling such inconsistent labels into 3D space leads to granularity blurring, where the model fails to assign a stable hierarchical rank to the same physical entity.

\textbf{Semantic-Scale Gap in Scale-based Distillation.}
To ensure view invariance, a second category of work \cite{kerr2023lerf,kim2024garfield,cen2025SAGA} groups 3D segments based on absolute physical scales. However, relying on a manual scale parameter to generate segments leads to semantic-scale decoupling, as semantic entities often exhibit high intra-class scale variance. For instance, as shown in~\cref{fig:intro_compare}(b), flowers within a bouquet vary greatly in size. Under a chosen scale parameter, a large flower might be over-segmented into petals, while a smaller flower at the same semantic level remains intact. Consequently, these methods require exhaustive per-instance tuning rather than discovering a consistent semantic hierarchy, hindering the automated reasoning essential for open-world understanding.

\textbf{Our Insight: From Granularities and Scales to Structural Levels.}
To build an accurate 3D semantic hierarchy, we argue that the system must shift from view-dependent \textit{granularities} and manual physical \textit{scales} to intrinsic \textbf{structural levels}—a consistent compositional rank that is strictly invariant to both viewing distances and absolute physical sizes (\eg, all flowers in a bouquet belong to the "bud" level, regardless of how large they are or how close the camera is).

To achieve this, we present LEGO (\textbf{LE}veled \textbf{G}aussian splatting for \textbf{O}pen-vocabulary understanding). 
Our fundamental insight is that while SAM’s 2D granularity labels are volatile across viewpoints, they collectively serve as partial glimpses into a scene's underlying semantic hierarchy. For example, a distal viewpoint might perceive "flowerpot $\to$ bouquet $\to$ bud" as its three granularities, while a proximal viewpoint captures "bouquet $\to$ bud $\to$ petal". Although their relative definitions shift, these observations jointly encapsulate the complete semantic lineage: "flowerpot $\to$ bouquet $\to$ bud $\to$ petal". The core of LEGO lies in its ability to adaptively re-grade all multi-view masks covering the same 3D region into a unified, level-wise structure, effectively "stitching" fragmented 2D granularities into a coherent 3D hierarchy. 

Specifically, we aggregate multi-view SAM masks and estimate their corresponding 3D physical scales. Since masks representing the same semantic entity exhibit consistent spatial dimensions across viewpoints, we utilize these physical scales as an \textit{initial proxy} to establish the semantic hierarchy. By clustering masks based on spatial co-visibility and 3D scale, we effectively discretize the continuous scales into distinct structural levels, thereby manifesting the implicit 3D hierarchy of the scene. 
As illustrated in~\cref{fig:intro_compare}(c), our 3D \textit{level-based} paradigm provides accurate, natively consistent supervision for hierarchical Gaussian scene segmentation, effortlessly parsing the scene into stable tiers.

To this end, we ground these segments with optimal-view CLIP features, establishing a \textit{multi-level, open-vocabulary hierarchy} for scene objects. 
We also show that, by further integrating spatial context of objects, we elevate these segments into \textit{level-wise language scene graphs}, effectively empowering Large Language Models (LLMs) to perform complex, context-aware spatial reasoning.

\section{Related Work}
\label{sec:related_work}

\subsection{Segmentation in Radiance Fields}
The emergence of Neural Radiance Fields (NeRF) \cite{mildenhall2021nerf} and 3D Gaussian Splatting (3DGS) \cite{kerbl20233dgs} has revolutionized 3D scene representation. While many methods \cite{zhu2024pcflift,ye2024gsgrouping,lyu2024gaga,yang2025instascene,zhu2025unifiedlift} successfully lift 2D instance masks to 3D Radiance Field, they predominantly focus on object-centric tasks, overlooking fine-grained part-level details essential for comprehensive scene understanding.

To capture these finer details, some works have ventured into part-level segmentation. However, addressing the inherent ambiguity of part decomposition remains a challenge. One line of work \cite{ying2024omniseg3d,kim2024garfield,cen2025SAGA} introduces global regulatory parameters—such as physical scale or feature similarity thresholds—to resolve this ambiguity. Although capable of separating parts, these methods often require manual tuning of these global parameters to match the specific granularity of a target object, limiting their ability to handle complex scenes with diverse intra-class scales. Another stream of research \cite{jain2024gaussiancut,zhang2025cobgs,zhao2025isegman} adopts a spatial interactive paradigm, relying on user clicks or 2D bounding boxes as initialization cues. While effective for human-in-the-loop applications, these methods lack the autonomy required for scalable, automatic processing. In contrast, LEGO achieves native, autonomous hierarchical segmentation by discovering consistent structural levels without predefined scale/similarity priors or manual guidance.

\subsection{Open-Vocabulary 3D Understanding}
To equip 3D representations with semantic understanding, early methods distilled patch-level CLIP embeddings \cite{kerr2023lerf,zuo2025fmgs,shi2024legaussian,liu20233dovs} or leveraged dense features from other 2D vision encoders \cite{peng2023openscene,qu2024goi,li2022lseg,ghiasi2022openseg,shen2024ape,caron2021dino,oquab2023dinov2}, but often struggled with blurred feature boundaries or long-tail categories.

With the advent of the Segment Anything Model (SAM), many works \cite{ye2024gsgrouping,ji2025fastlgs,peng20243dvlgs,wu2024opengaussian,li2025instancegaussian,jun2025drsplat,zhu2025cos3d,zhu2026eps3d} utilize SAM masks to aggregate CLIP features, significantly improving boundary adherence. However, these methods primarily operate at the object level. While some attempts \cite{qin2024langsplat,peng2024gags,cen2025laga,cheng2024occamlgs,dai2025thgs} have explored part-level lifting, they often encounter severe context deficiency. This is particularly problematic for low-texture components, which lack distinctive visual features when isolated from the whole object, leading to incorrect feature extraction. To mitigate this, several approaches \cite{bhalgat2024n2f2,sun2025cags,huang2025openinsgaussian,tan2025fmlgs} tried to incorporate context by fusing features from surrounding or parent objects. Nevertheless, this strategy inevitably leads to feature averaging and blurring, resulting in semantic ambiguity where distinct parts become indistinguishable from their neighbors. Diverging from these, LEGO employs a structurally decoupled feature space. By optimizing hierarchical levels independently, we strictly prevent cross-level semantic entanglement, extracting pristine features for robust, context-aware part-level understanding.

\subsection{Structured Scene Modeling}
Beyond dense semantic fields, structuring 3D scenes into graphs provides a more interpretable representation. Object-level scene graphs \cite{gu2024conceptgraphs,wang2025gaussiangraph,wang2025ov_octreegraph,linok2025bbq} abstract scenes into nodes representing objects and edges representing spatial relationships. While effective for navigation, they lack the granularity to describe intra-object structures. To incorporate hierarchy, HOV-SG \cite{werby2024hov-sg} constructs a multi-layer floor-room-object graph but stops short of decomposing objects into parts. Recently, a few works \cite{dai2025thgs,zhan2025hilsplat} have attempted part-level hierarchical modeling. However, these methods are fundamentally constrained by SAM’s architecture, which only provides three predefined output levels (whole, part, subpart). This rigid limitation prevents them from adapting to the varied complexity of real-world scenes. LEGO transcends this limitation by proposing an adaptive decomposition strategy. Unconfined by view-dependent 2D granularities, LEGO constructs a flexible, multi-layer 3D scene graph, effectively empowering Large Language Models (LLMs) to perform complex, context-aware spatial reasoning.

\section{Method}

Given a set of multi-view RGB images $I=\{I_1,I_2,\dots,I_V\}$ of a scene, LEGO aims to output a 3D Gaussian field enriched with hierarchical semantic features, allowing a multi-level scene graph construction for LLM-based scene reasoning.

Specifically, LEGO first initializes a 3D Gaussian field \cite{wang2025vistadream} $\mathcal{P}$ using the geometric prior reconstructed by MASt3R-SfM \cite{duisterhof2025mast3r-sfm}. 
Then, in \cref{sec:3.1}, we re-grade multi-view SAM masks into 3D-consistent leveled masks. These masks then facilitate the hierarchical segmentation of $\mathcal{P}$ in \cref{sec:3.2}. Finally, in \cref{sec:3.3}, we embed CLIP language features into the resulting segments and organize them into a scene graph based on their spatial relationships, empowering Large Language Models to perform complex spatial grounding tasks.

\begin{figure}[tb]
  \centering
  \includegraphics[width=\linewidth]{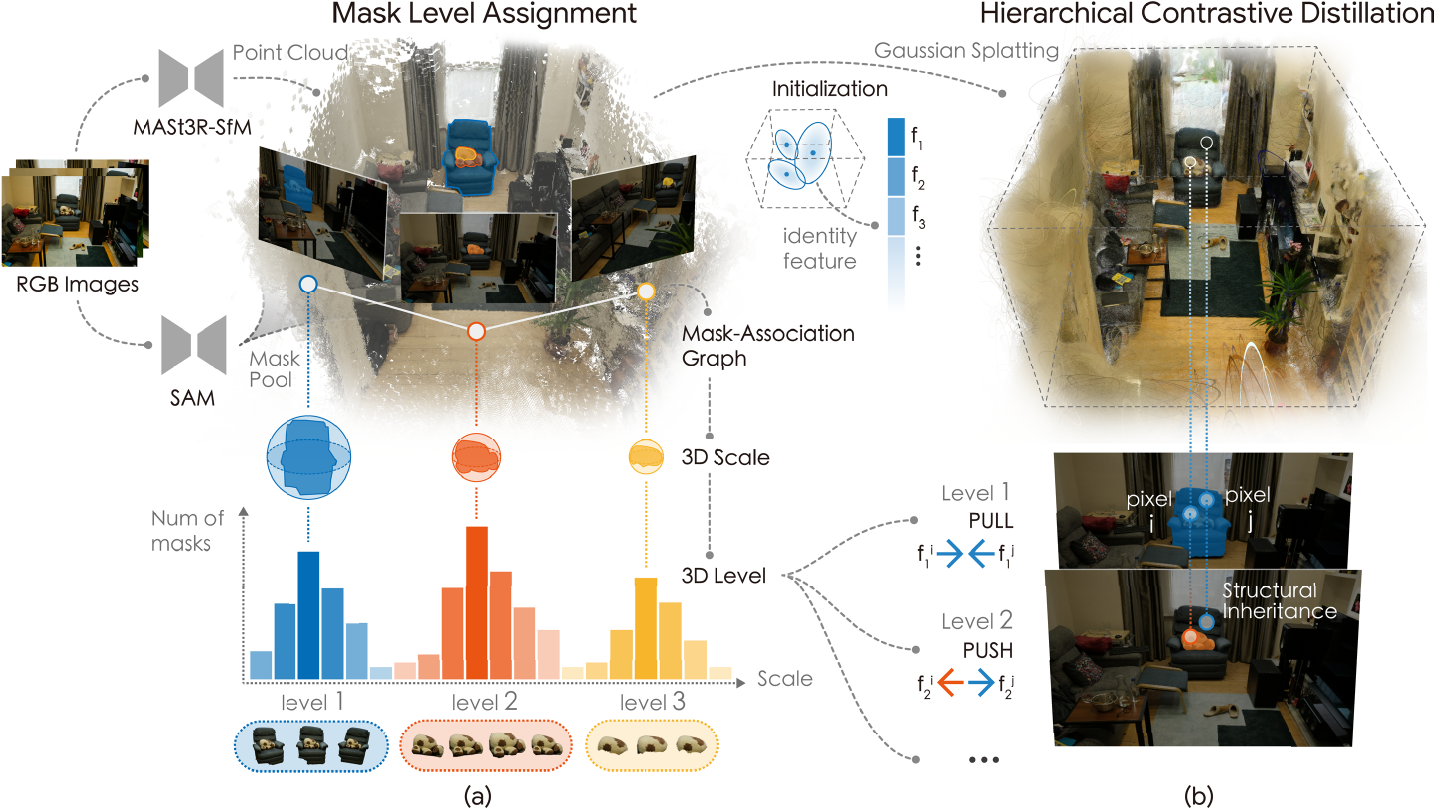}
  \caption{\textbf{LEGO pipeline.} 
  \textbf{(a)} LEGO first lifts multiview SAM masks into a 3D coherent scale space to assign them view-consistent structural levels (\cref{sec:3.1}). \textbf{(b)} Guided by these level-masks, a hierarchical contrastive distillation process optimizes the decoupled identity features of the Gaussian field (\cref{sec:3.2}).}
  \label{fig:pipeline}
\end{figure}

\subsection{3D Semantic Hierarchy from SAM Masks}
\label{sec:3.1}
In this section, we re-grade multi-view, multi-granularity SAM masks into a consistent 3D semantic hierarchy. Specifically, each mask is assigned a structural level label that characterizes its relative semantic scale within its local region.

\textbf{Mask Lifting and 3D Scale Estimation.} We first aggregate multi-view SAM masks into a global pool $\mathcal{M}$. Using camera poses and scene representation $\mathcal{P}$ from MASt3R-SfM \cite{duisterhof2025mast3r-sfm}, we establish a pixel-to-point mapping $\mathcal{F}: u \mapsto \mathbf{p}$. For each 2D mask $m_i \in \mathcal{M}$, we lift its constituent pixels to 3D points $\mathcal{P}_{m_i} = \{ \mathcal{F}(u) \mid u \in m_i \}$. The physical scale $s_i$ of mask $m_i$ is derived as the effective spatial diameter based on the point set's standard deviation: $s_i = 2\sqrt{\sum_{d \in \{x,y,z\}} \operatorname{std}(\mathcal{P}_{m_i,d})^2}$.


\textbf{Local Peak-based Mask Level Assignment.}
We determine the hierarchical level of $m_i$ by situating it within its local structural context. Specifically, for each target mask $m_i$, we first identify its spatially neighboring masks $\mathcal{N}(m_i)$ that share co-visible 3D regions with it. 

\begin{figure}[tb]
  \centering
  \includegraphics[width=\linewidth]{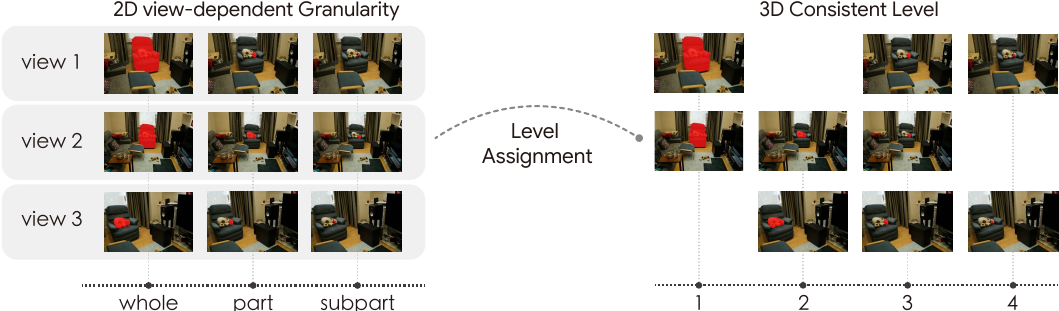}
  \caption{
  \textbf{Illustration of mask level assignment.} Our mechanism discovers the complete 3D hierarchical structure implicit in multi-view 2D masks, assigning each mask to its corresponding view-consistent level.
  }
  \label{fig:pipeline_level_assign}
\end{figure}

Then, we perform peak detection on the histogram of $\mathcal{N}(m_i) \cup \{m_i\}$ scales to self-adaptively identify $L$ prominent peaks $\mathcal{K}_i=\{p_1, p_2, \dots, p_L\}$, sorted by scale from coarse to fine.
These peaks represent the underlying semantic hierarchy (\eg, $p_1$ for the root object, $p_L$ for the finest sub-parts) present in this local 3D region.
Finally, the target mask $m_i$ is assigned a discrete level $l_i$ based on which peak its own scale $s_i$ aligns with:
\begin{equation}
    l_i = \mathop{\arg\min}_{l \in \{1, 2, \dots, L\}} |s_i - p_l|.
\end{equation}
Through this per-mask assignment, as illustrated in \cref{fig:pipeline_level_assign}, LEGO ensures that each mask is "self-graded" into a view-consistent local 3D semantic hierarchy, effectively resolving the ambiguity of 2D granularity by leveraging the collective consensus of its co-visible neighbors.

\textbf{Level-wise Mask Dense Indicator.}
Once each mask is assigned to a structural level, a naive supervision strategy would be to optimize level-$k$ segmentation using only the masks strictly graded at that level. 
However, because our reconstructed 3D hierarchy contains significantly more levels than SAM's three 2D granularities, the re-graded SAM masks for any single level are inherently sparse. Supervision solely with these sparse masks leads to optimization instability and structural artifacts in level-wise 3D segmentation. 
To resolve this, we design several strategies to densify the level-wise supervision by propagating mask labels through the hierarchy. 

In implementation, as illustrated in \cref{fig:pipeline}(b), rather than explicitly densifying levelwise mask maps, we derive an equivalent but highly parallelizable pixel-wise indicator $\mathbb{I}$: for any pixel pair $(i, j)$ in view $v$, $\mathbb{I}_k(i, j)$ determines whether they belong to the same entity mask at level $k$, formulated as 
\begin{equation}
\mathbb{I}_k(i, j) =
\begin{cases}
1 & \text{if } \mathcal{M}_k(i) \cap \mathcal{M}_k(j) \neq \emptyset \\
0 & \text{otherwise},
\end{cases}
\end{equation}
where $\mathcal{M}_k(i)$ denotes the set of active masks covering pixel $i$ at level $k$.

Notably, the construction of $\mathbb{I}$ is non-trivial, requiring strict adherence to the following hierarchical axioms to maintain structural integrity:
\begin{itemize}
    \item \textbf{Monotonicity}: Supervisory signals are strictly non-increasing as the hierarchy deepens; pixel pairs may split into sub-parts as granularity increases, but they can never merge at finer levels.
    \item \textbf{Recursive Inclusion}: Identity at a fine-grained level strictly necessitates identity at all coarser parent levels, ensuring nested structural consistency across the field.
    \item \textbf{Structural Inheritance}: To mitigate mask sparsity and prevent optimization collapse, pixel lacking a valid mask at the current target level inherit labels from their nearest valid coarser level, ensuring dense and stable supervision at every layer.
\end{itemize}

These principles yield a conflict-free and hierarchically-consistent dense supervision by pixel-pair parallelism, stabilizing efficient level-wise segmentation in the following section. 
More details are provided in the Supplementary Material.

\subsection{Segmentation with Level-wise Semantic Hierarchy}
\label{sec:3.2}
In this section, we describe the supervision of the hierarchical segmentation for Gaussian field $\mathcal{P}$ using the level-wise topological constraints established above. Inspired by SAGA~\cite{cen2025SAGA}, we equip each Gaussian primitive with a learnable identity embedding. By enforcing the rendered feature distributions to align with our level-wise indicator constraints, the model learns to capture the spatial and semantic distinctness of different regions. To support this hierarchical supervision, we embed a decoupled feature space where each structural level is optimized independently without cross-level interference.

\textbf{Level Feature Embedding.} To represent a multi-granular hierarchy without inter-level interference, we assign each Gaussian primitive in $\mathcal{P}$ a composite identity feature $\mathbf{f} \in \mathbb{R}^{L \times d}$. We structurally decouple the feature space into $L$ independent $d$-dimensional subspaces, where the feature for level $k$ is directly routed to its $k$-th row vector $\mathbf{f}_k \in \mathbb{R}^d$. This disjoint design prevents semantic entanglement, allowing the model to optimize the feature representation of each hierarchical level independently. 

\textbf{Level-wise Feature Learning}. 
We formulate a contrastive distillation loss to dynamically align the Gaussian features with the dense hierarchical supervision. For a given training viewpoint $v$ and target level $k$, we first render the level-specific primitive features $\mathbf{f}_k$ to a 2D feature map $\mathbf{F}_{v,k}$. 

For a sampled pixel pair $(i, j)$ at level $k$, the contrastive objective is defined as:
\begin{equation}
\mathcal{L}_{contrast} = w^{i,j}_k \left( 1 - 2 \cdot \mathbb{I}_k(i, j) \right) \cos(\mathbf{F}_{v,k}^i, \mathbf{F}_{v,k}^j) 
\end{equation}
where $\mathbf{F}_{v,k}^i$ is the rendering feature of pixel $i$. The indicator function $\mathbb{I}_k(i, j)$, as formulated in \cref{sec:3.1}, dynamically dictates the pairing relationship based on the intersection of the active mask sets $\mathcal{M}_k(i)$ and $\mathcal{M}_k(j)$. 

To prevent large-scale entities from dominating the gradient, we employ a scale-balancing weight $w^{i,j}_k \propto (\mathcal{A}^i_k \cdot \mathcal{A}^j_k)^{-1}$, where $\mathcal{A}^i_k$ denotes the area of the active mask governing pixel $i$ at level $k$. This ensures that small objects contribute equally to the optimization as massive ones within each level.
Furthermore, as cosine similarity only constrains feature orientation, we impose an auxiliary $L_2$ penalty on positive pairs to enforce strict feature identity:
\begin{equation}
\mathcal{L}_{pos} = \mathbb{I}_k(i, j) \|\mathbf{F}_{v,k}^i - \mathbf{F}_{v,k}^j\|_2^2
\end{equation}

Moreover, since rendered pixels are weighted accumulations of multiple Gaussians, the model might exploit view-dependent shortcuts that fail to represent consistent 3D geometry. We counteract this by constraining both 3D primitive features $\mathbf{f}_k$ and rendered 2D features $\mathbf{F}_{v,k}^i$ to a unit hypersphere:
\begin{equation}
\mathcal{L}_{3d} = |1 - \|\mathbf{f}_k\|_2|, \quad \mathcal{L}_{2d} = |1 - \|\mathbf{F}_{v,k}^i\|_2|
\end{equation}
This ensures that all Gaussians along a single ray possess identical, consistent features, preventing feature decay during blending for a truly view-invariant 3D semantic field.

\textbf{Hierarchical Scene Segmentation.}
After convergence of the above feature learning, LEGO performs hierarchical segmentation through a top-down recursive clustering strategy, naturally representing the scene as a tree-structured semantic hierarchy. 
Starting from the root node (the entire scene), we iteratively partition parent nodes into child clusters. Specifically, for a parent node at level $k-1$, we apply HDBSCAN \cite{mcinnes2017hdbscan} to its constituent Gaussian primitives based on their level-$k$ features $\mathbf{f}_k$. This process continues recursively until the maximum hierarchy depth $L$ is reached. By leveraging the decoupled nature of our feature subspaces, this approach ensures that the resulting segments are not merely isolated clusters; instead, they are structurally nested to faithfully reflect the multi-granular semantic hierarchy distilled from the leveled masks.

\subsection{Level-wise Language Scene Graph}
\label{sec:3.3}
In this section, we endow the hierarchical segments with CLIP embeddings for open-vocabulary understanding. To further facilitate complex spatial reasoning, we incorporate spatial context to synthesize these language-grounded segments into level-wise scene graphs. This structured representation captures both hierarchical containment (part-whole) and spatial proximity, effectively empowering the model to resolve complex, context-aware queries such as "Find the handle of the pitcher beside the rolling pin."

\textbf{View-Aware Language Grounding.}
When embedding image CLIP features into a 3D segment, to avoid semantic noise introduced by occlusions in a naive multi-view average, we propose an Optimal View Selection (OVS) strategy. For segment $c$, the observation quality score $\mathcal{S}(c, v)$ for view $I_v$ is:
\begin{equation}
    \mathcal{S}(c, v) = \underbrace{ \left( \frac{|c_{\text{vis}}^v|}{|{c}|} \right) }_{\text{3D Visibility}} \cdot \underbrace{ \left( \frac{|m_c^v|}{|H \times W|} \right) }_{\text{2D Coverage}} \cdot \underbrace{ \text{IoU}(m_c^v, m^*) }_{\text{Semantic Alignment}}
\end{equation}

where $|{c}|$ denotes the total number of Gaussians in $c$, $|{c}_{\text{vis}}^v|$ is the subset of those Gaussians successfully mapped to valid pixels in frame $I_v$, $m_c^v$ is the projected 2D mask of $c$, $m^*$ represents the best-aligned SAM mask in view $v$, and $H \times W$ is the total pixel area of the image. "3D Visibility" penalizes segments that are heavily occluded in the current frame, "2D Coverage" prioritizes viewpoints where the projected segment occupies a significant portion of the view, "Semantic Alignment" ensures the projected segment mask $m_c^v$ aligns well with the 2D SAM prior $m^*$.

Then, we retrieve the top-$\tau$ frames that maximize $\mathcal{S}(c, v)$. The corresponding high-quality SAM masks $m^*$ are utilized to crop the input images and extract CLIP embeddings. The final open-vocabulary feature $\mathbf{E}_c$ is obtained via average pooling of these top-$\tau$ embeddings. This view-aware strategy ensures pristine, high-fidelity semantic representations while drastically reducing computational overhead.

\textbf{From Segments to Scene Graph.}
To move beyond a simple semantic tree and capture the full spatial context, we define two types of edges:
\begin{itemize}
    \item Hierarchical Edges ($\mathcal{E}_{hier}$): These encode the vertical "part-whole" relationships established during the hierarchical scene segmentation process.
    \item Spatial Adjacency Edges ($\mathcal{E}_{adj}$): These represent the horizontal context. We establish an edge between two independent nodes if their 3D bounding spheres intersect. This provides the topological foundation for reasoning about spatial relations like "on" or "beside".
\end{itemize}

The resulting scene graph enables sophisticated tasks like Chain-of-Retrieval (CoR). When faced with a complex query (\eg, "Find the handle of the pitcher beside the rolling pin"), an LLM agent parses it into a coarse-to-fine sequence (rolling pin $\to$ pitcher $\to$ handle). LEGO then executes a search over the graph topology, using coarser objects as spatial anchors to retrieve fine-grained parts through semantic matching and spatial relationship verification.
\begin{figure}[tb]
  \centering
  \includegraphics[width=\linewidth]{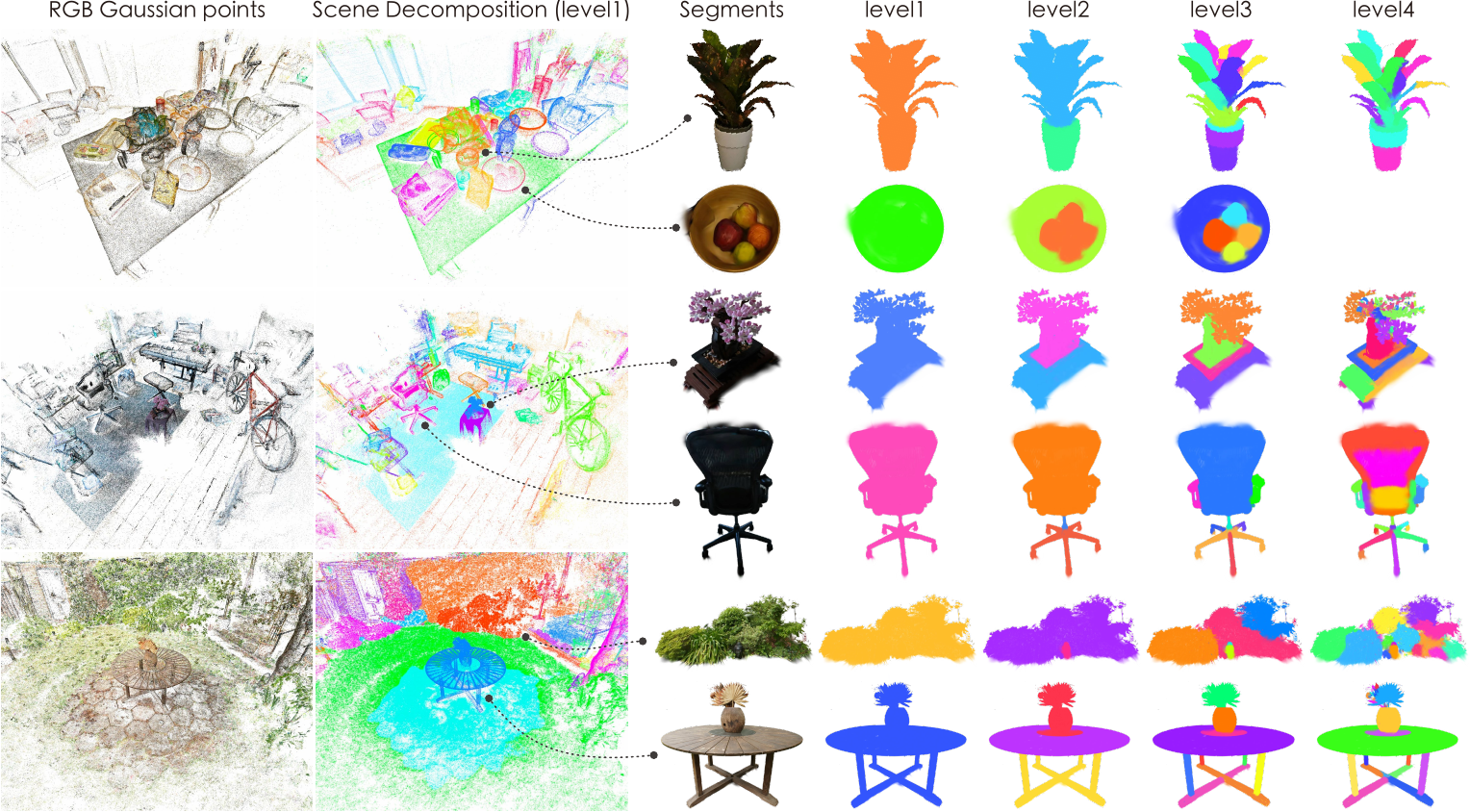}
  \caption{\textbf{Visualization of automatic hierarchical scene decomposition. } 
  Benefiting from our level-wise design, LEGO captures arbitrarily deep hierarchical structures. Here, selected entities are progressively parsed into extreme fine-grained sub-parts, while maintaining strict structural nestedness across multiple levels.
  }
  \label{fig:decompose_vis_main}
\end{figure}

\section{Experiments}

\textbf{Tasks and Datasets.}
We comprehensively evaluate the performance of LEGO on both segmentation and understanding tasks:

\textbf{Promptable Segmentation.} We conduct experiments on the NVOS \cite{ren2022nvos} and SPIn-NeRF \cite{mirzaei2023spinnerf} datasets. The NVOS dataset, derived from the established LLFF \cite{mildenhall2019llff} dataset, comprises 7 diverse real-world scenes with complex geometric structures. The SPIn-NeRF dataset integrates subsets from several existing NeRF benchmarks, providing high-quality real-world captures designed to evaluate precise multi-view segmentation.

\textbf{Open-vocabulary Understanding.} We utilize two real-world datasets: the LERF-OVS dataset \cite{kerr2023lerf} (extended with LangSplat \cite{qin2024langsplat} annotations) and Mip-NeRF 360 dataset \cite{barron2022mipnerf360} (annotated by GAGS \cite{peng2024gags}). The LERF dataset contains over 3,700 high-resolution images captured by mobile phones across 14 diverse scenes. LangSplat provides text descriptions and corresponding multi-view segmentation masks within four scenes (\textit{``ramen''}, \textit{``waldo\_kitchen''}, \textit{``teatime''}, and \textit{``figures''}) to assess text-based localization and segmentation. The Mip-NeRF 360 dataset offers unbounded and highly intricate indoor and outdoor scenes. Following GAGS, we evaluate text-based localization and segmentation in four scenes (\textit{``room''}, \textit{``counter''}, \textit{``garden''}, and \textit{``bonsai''}).

For open-vocabulary evaluations with single-word or phrase queries, we bypass the LLM scene graph (\cref{sec:3.3}) and simply perform standard CLIP feature similarity across the first three hierarchical levels to retrieve the target object.

\textbf{Evaluation Metrics.}
We evaluate promptable segmentation using standard mean Intersection over Union (mIoU) and mean Accuracy (mAcc). For open-vocabulary tasks, we assess semantic grounding via localization mAcc (predictions within ground-truth bounding boxes) and text-queried segmentation mIoU.

\begin{table}[tb]
  \begin{minipage}{0.5\textwidth}
    \centering
    \fontsize{8pt}{9pt}\selectfont
    \caption{Results on NVOS dataset.}
    \label{tab:result_NVOS}
    \begin{tabular}{l|cc}
        \toprule
        Method & mIoU(\%) & mAcc(\%)\\
        \midrule
        NVOS \cite{ren2022nvos} & 70.1 & 92.0 \\
        ISRF \cite{goel2023isrf} & 83.8 & 96.4 \\
        SA3D \cite{cen2023sa3d-nerf} & 90.3 & 98.2 \\
        OmniSeg3D \cite{ying2024omniseg3d} & 91.7 & 98.4 \\
        \midrule
        GS-Grouping \cite{ye2024gsgrouping} & 85.6 & 97.3 \\
        FlashSplat \cite{shen2024flashsplat} & 91.8 & \underline{98.6} \\
        iSegMan \cite{zhao2025isegman} & 92.0 & 98.4 \\
        COB-GS \cite{zhang2025cobgs} & 92.1 & \underline{98.6} \\ 
        SA3D-GS \cite{cen2025sa3d} & 92.2 & 98.5 \\
        SAGA \cite{cen2025SAGA} & \underline{92.6} & \underline{98.6} \\
        \midrule
        \textbf{LEGO(Ours)} & \textbf{94.2} & \textbf{98.7} \\
        \bottomrule
    \end{tabular}
  \end{minipage}\hfill
  \begin{minipage}{0.5\textwidth}
    \centering
    \fontsize{8pt}{9pt}\selectfont
    \caption{Results on SPIn-NeRF dataset.}
    \label{tab:result_SPinNeRF}
    \begin{tabular}{l|cc}
        \toprule
        Method & mIoU(\%) & mAcc(\%)\\
        \midrule
        MVSeg \cite{mirzaei2023spinnerf} & 90.9 & 98.9 \\
        SA3D \cite{cen2023sa3d-nerf} & 92.4 & 98.9 \\
        OmniSeg3D \cite{ying2024omniseg3d} & \textbf{94.3} & \textbf{99.3} \\
        \midrule
        GS-Grouping \cite{ye2024gsgrouping} & 86.5 & 98.9 \\
        SAGD \cite{hu2024sagd} & 90.0 & 98.7 \\
        iSegMan \cite{zhao2025isegman} & 92.4 & 99.1 \\ 
        SA3D-GS \cite{cen2025sa3d} & 93.2 & 99.1 \\
        SAGA \cite{cen2025SAGA} & 93.4 & \underline{99.2} \\
        \midrule
        \textbf{LEGO(Ours)} & \underline{94.2} & \textbf{99.3} \\
        \bottomrule
    \end{tabular}
  \end{minipage}
\end{table}

\subsection{Results}

\textbf{Promptable Segmentation.} \Cref{tab:result_NVOS,tab:result_SPinNeRF} report quantitative results on NVOS and SPIn-NeRF. LEGO achieves superior or highly competitive performance compared to prior methods, notably securing non-trivial margins (up to +1.6\% mIoU on NVOS) on these heavily saturated benchmarks. By explicitly grouping masks into distinct hierarchical levels and optimizing disjoint sub-features, LEGO insulates the representation from the noise of low-quality priors. This eliminates cross-level feature entanglement, resulting in a purer and strictly consistent feature field at every level.

\textbf{Open-Vocabulary Understanding.} As shown in \Cref{tab:result_LERF,tab:result_360}, LEGO consistently achieves state-of-the-art performance, yielding average improvements of +4.1\% localization mAcc and +4.4\% segmentation mIoU on LERF-OVS, alongside +3.9\% mAcc and +3.6\% mIoU on Mip-NeRF 360, achieving top or near-top performance across the vast majority of scenes.

\begin{figure}[ht]
  \centering
  \includegraphics[width=0.96\linewidth]{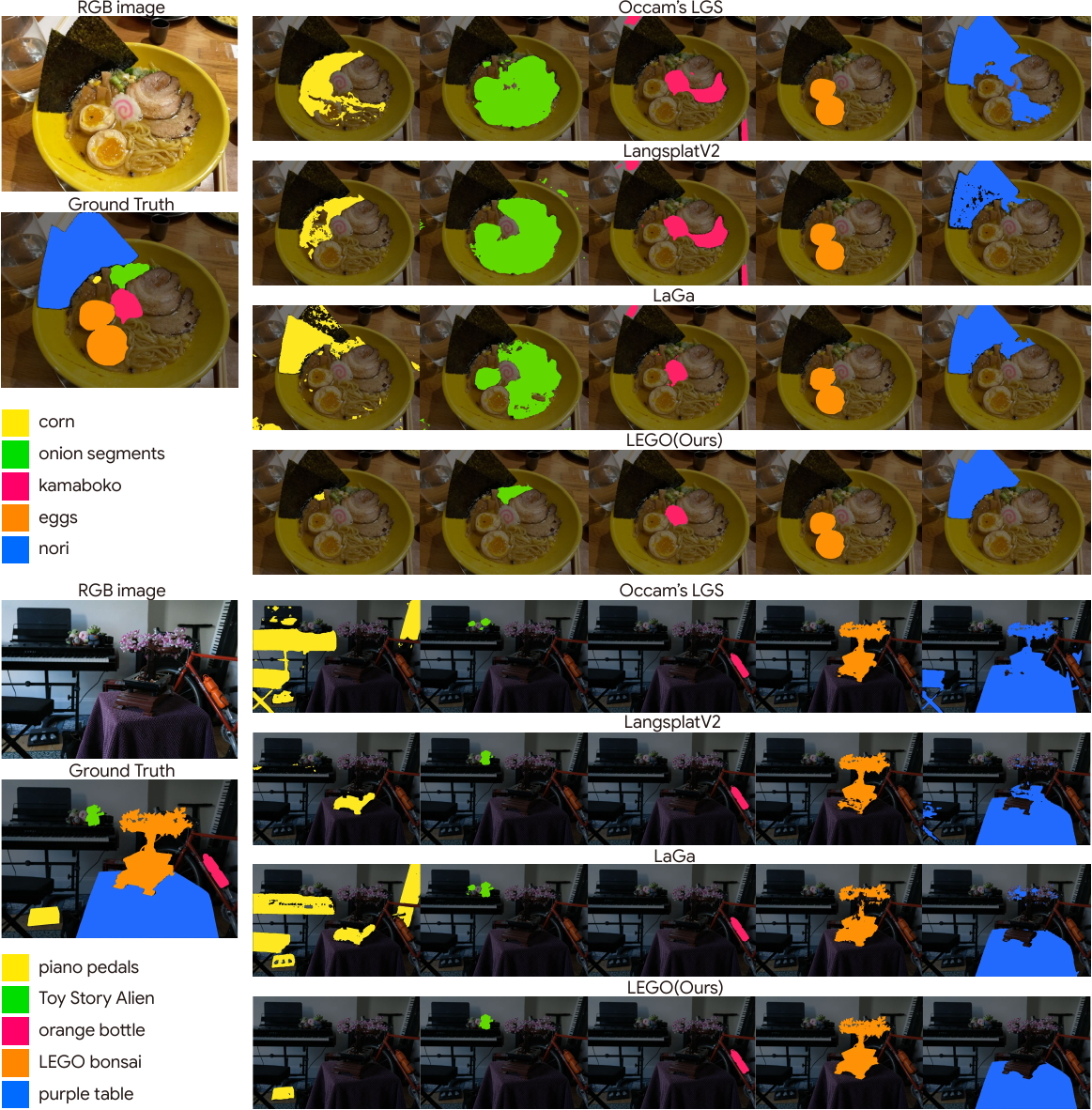}
  \caption{\textbf{Open-vocabulary segmentation results.} While other methods struggle with severe feature blurring, LEGO cleanly isolates extremely fine-grained entities like \textit{``corn''} and \textit{``onion segments''}, and preserves exquisite boundaries for \textit{``LEGO bonsai''}.
  }
  \label{fig:ovs_vis_main_1}
\end{figure}

Crucially, LEGO excels in segmenting extreme fine-grained sub-parts. In \cref{fig:ovs_vis_main_1}, while baselines only locate prominent entities like \textit{``eggs''} in the \textit{Ramen} scene, LEGO cleanly isolates fine-grained \textit{``corn''} and \textit{``onion segments''}, yielding a +11.2\% mAcc and +11.9\% mIoU boost over the best baseline. Similarly, in the \textit{Bonsai} scene, LEGO successfully segments tiny objects (\eg, \textit{``pedal of piano''}) and preserves exquisite boundaries for topologically complex objects with intricate edges (\eg, \textit{``LEGO bonsai''}), whereas alternative methods fail or erroneously highlight the entire parent object due to feature blending.

\begin{table}[tb]
  \caption{Open-vocabulary 3D understanding results on LERF-OVS dataset.}
  \label{tab:result_LERF}
  \centering
  \resizebox{\textwidth}{!}{
  \begin{tabular}{l|ccccc|ccccc}
    \toprule
    \multirow{2}{*}{Method} & \multicolumn{5}{c|}{3D Location mAcc(\%)} & \multicolumn{5}{c}{3D Segmentation mIoU(\%)} \\
    & \textit{Ramen} & \textit{Teatime} & \textit{Kitchen} & \textit{Figurines} & \textit{\textbf{Overall}} & \textit{Ramen} & \textit{Teatime} & \textit{Kitchen} & \textit{Figurines} & \textit{\textbf{Overall}} \\
    \midrule
    LEGaussian \cite{shi2024legaussian} & 69.0 & 79.7 & 63.6 & 57.1 & 67.4 & 20.2 & 32.3 & 22.3 & 23.4 & 24.6 \\
    GS-Grouping \cite{ye2024gsgrouping} & 32.4 & 69.5 & 50.0 & 44.6 & 49.1 & 26.4 & 54.0 & 31.3 & 34.6 & 36.6 \\ 
    GOI \cite{qu2024goi} & 56.3 & 67.8 & 68.2 & 44.6 & 59.2 & 33.7 & 55.8 & 54.5 & 23.9 & 42.0 \\
    GAGS \cite{peng2024gags} & 69.0 & 88.1 & \underline{90.9} & 78.6 & 81.7 & 46.8 & 60.3 & 55.8 & 53.6 & 54.1 \\
    LangSplat \cite{qin2024langsplat} & 73.2 & 88.1 & \textbf{95.5} & \underline{80.4} & \underline{84.3} & 51.2 & 65.1 & 44.5 & 44.7 & 51.4 \\
    Occam’s LGS \cite{cheng2024occamlgs} & 74.7 & \underline{93.2} & 81.8 & \underline{80.4} & 82.5 & 51.0 & 70.2 & \underline{65.3} & 58.6 & 61.3 \\
    LangSplatV2 \cite{li2025langsplatv2} & 74.7 & \underline{93.2} & 86.4 & \textbf{82.1} & 84.1 & 51.8 & \underline{72.2} & 59.1 & 56.4 & 59.9 \\
    LaGa \cite{cen2025laga} & \underline{76.1} & 84.8 & 81.8 & 78.6 & 80.3 & \underline{55.6} & 70.9 & \textbf{65.6} & \textbf{64.1} & \underline{64.0} \\
    \midrule
    \textbf{LEGO(Ours)} & \textbf{87.3} & \textbf{94.9} & \underline{90.9} & \underline{80.4} & \textbf{88.4} & \textbf{67.5} & \textbf{81.0} & 62.7 & \underline{62.3} & \textbf{68.4} \\

  \bottomrule
  \end{tabular}
  }
\end{table}

\begin{table}[tb]
  \caption{Open-vocabulary 3D understanding results on Mip-NeRF 360 dataset.}
  \label{tab:result_360}
  \centering
  \resizebox{\textwidth}{!}{
  \begin{tabular}{l|ccccc|ccccc}
    \toprule
    \multirow{2}{*}{Method} & \multicolumn{5}{c|}{3D Location mAcc(\%)} & \multicolumn{5}{c}{3D Segmentation mIoU(\%)} \\
    & \textit{Room} & \textit{Counter} & \textit{Garden} & \textit{Bonsai} & \textit{\textbf{Overall}} & \textit{Room} & \textit{Counter} & \textit{Garden} & \textit{Bonsai} & \textit{\textbf{Overall}} \\
    \midrule  
    LEGaussian \cite{shi2024legaussian} & 55.2 & 73.0 & 71.4 & 61.1 & 65.2 & 25.5 & 35.3 & 33.2 & 22.3 & 29.1 \\ 
    GS-Grouping \cite{ye2024gsgrouping} & \underline{79.3} & 56.8 & 57.1 & 66.7 & 65.0 & 54.4 & 47.7 & 40.4 & 54.1 & 49.1 \\
    GOI \cite{qu2024goi} & 69.0 & 67.6 & 66.7 & 72.2 & 68.9 & 60.3 & 46.6 & 59.8 & 67.3 & 58.5 \\
    GAGS \cite{peng2024gags} & \textbf{93.1} & \textbf{97.3} & \underline{81.0} & \underline{83.3} & \underline{88.7} & 65.2 & 61.1 & 61.2 & 70.5 & 64.5 \\
    LangSplat \cite{qin2024langsplat} & 75.9 & 91.9 & 52.4 & 72.2 & 73.1 & 51.2 & 64.7 & 49.8 & 53.0 & 54.7 \\
    Occam’s LGS \cite{cheng2024occamlgs} & \textbf{93.1} & \textbf{97.3} & 57.1 & 72.2 & 79.9 & \underline{69.2} & \underline{72.9} & 50.9 & 62.2 & 63.8 \\
    LangSplatV2 \cite{li2025langsplatv2} & 75.9 & 91.9 & \textbf{85.7} & \underline{83.3} & 84.2 & 64.3 & \textbf{75.1} & \underline{65.0} & \underline{73.1} & \underline{69.4} \\
    LaGa \cite{cen2025laga} & \textbf{93.1} & \underline{94.6} & 71.4 & \underline{83.3} & 85.6 & \textbf{72.6} & 72.1 & 42.1 & 68.0 & 63.7 \\
    \midrule
    \textbf{LEGO(Ours)} & \textbf{93.1} & \textbf{97.3} & \textbf{85.7} & \textbf{94.4} & \textbf{92.6} & 67.3 & 71.3 & \textbf{71.0} & \textbf{82.4} & \textbf{73.0} \\

  \bottomrule
  \end{tabular}
  }
\end{table}

\begin{figure}[tb]
  \centering
  \includegraphics[width=\linewidth]{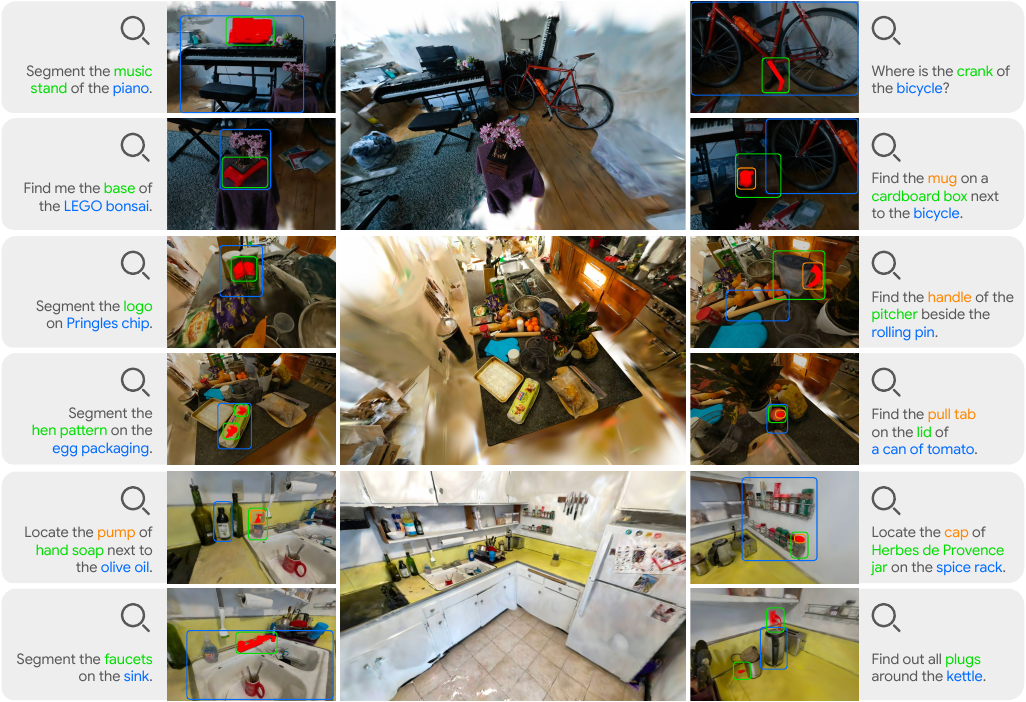}
  \caption{\textbf{Visualization of graph-based Chain-of-Retrieval.} Colored words match bounding boxes, illustrating how coarser anchors guide fine-grained localization.
  }
  \label{fig:ovs_vis_llm_query}
\end{figure}

\textbf{LLM-Guided Complex Scene Reasoning.}
We further evaluate LEGO's capability to parse compositional prompts that require deep hierarchical and relational reasoning—a known failure point for standard CLIP-based fields due to the "bag-of-words" effect. As illustrated in~\cref{fig:ovs_vis_llm_query}, we apply the LLM-guided Chain-of-Retrieval (CoR) over our constructed scene graph. For highly intricate spatial queries (\eg, \textit{``Find the mug on a cardboard box next to the bicycle''}) or extreme fine-grained part localization (\eg, \textit{``Segment the logo on Pringles chip''}), LEGO seamlessly handles them by utilizing coarser objects as spatial anchors to progressively filter context, successfully resolving semantic ambiguity and pinpointing tiny targets that require rich hierarchical and spatial understanding. 

To quantitatively evaluate this capability, we introduce a fine-grained CoR benchmark comprising 120 complex queries across four diverse scenes (30 queries per scene, analogous to the qualitative examples in~\cref{fig:ovs_vis_llm_query}). Each query entails intricate part-whole and spatial relationships, primarily targeting fine-grained elements. As reported in \cref{tab:CoR_metric}, we compare LEGO against scene graph methods (THGS \cite{dai2025thgs}, BBQ \cite{linok2025bbq}) and a strong baseline (LaGa). Note that LaGa is evaluated both in its native state and when augmented with our CoR strategy. BBQ and THGS construct scene graphs lacking semantic levels and consequently perform poorly. While adding a graph module to LaGa yields noticeable gains, its fixed, sparse SAM granularities (limited to 3 levels) suffer from multi-view inconsistency and fail to capture the accurate hierarchy essential for mixed relational querying. In contrast, LEGO maintains robust performance and achieves state-of-the-art results by unifying 3D-consistent semantic hierarchies with spatial relations.

\begin{table}[ht]
  \caption{Quantitative evaluation of CoR queries (mIoU, \%). We compare LEGO against scene graph baselines (BBQ, THGS) and a strong baseline LaGa, where LaGa is evaluated both natively and when augmented with our CoR strategy.}
  \label{tab:CoR_metric}
  \centering
  \begin{tabular}{lccccc}
    \toprule
    Method & \textit{Teatime} & \textit{Kitchen} & \textit{Bonsai} & \textit{Counter} & \textit{\textbf{Overall}} \\
    \midrule
    BBQ \cite{linok2025bbq} & 5.6 & 2.6 & 8.1 & 8.2 & 6.1 \\
    THGS \cite{dai2025thgs} & 9.9 & 5.0 & 10.3 & 12.9 & 9.5 \\
    LaGa \cite{cen2025laga} & 5.8 & 5.2 & 12.0 & 10.4 & 8.3 \\
    LaGa w/ CoR & 8.4 & 14.5 & 13.8 & 20.5 & 14.3 \\
    LEGO w/o CoR & 25.2 & 10.0 & 30.2 & 25.6 & 22.7 \\
    \textbf{LEGO(Ours)} & \textbf{52.2} & \textbf{44.9} & \textbf{57.6} & \textbf{51.4} & \textbf{51.6} \\
    \bottomrule
  \end{tabular}
\end{table}

\begin{table}[ht]
  \caption{Ablation studies on the \textit{Room} scene of the Mip-NeRF 360 dataset. We progressively analyze the impact of our scale-balancing re-weighting ($\mathcal{W}$), positive $L_2$ penalty ($\mathcal{L}_{pos}$), and feature normalization ($\mathcal{L}_{3d/2d}$).}
  \label{tab:ablation}
  \centering
  \begin{tabular}{ccc|cc}
    \toprule
    \multicolumn{3}{c|}{\textbf{Settings}} & \multicolumn{2}{c}{\textbf{Metrics}} \\
    Re-weighting ($\mathcal{W}$) & pos ($\mathcal{L}_{pos}$) & Norm ($\mathcal{L}_{3d/2d}$) & mAcc(\%) & mIoU(\%) \\
    \midrule  
    \checkmark & \checkmark & \checkmark & \textbf{93.1} & \textbf{67.3} \\
               & \checkmark & \checkmark & 86.2 & 63.6 \\
               &            & \checkmark & 86.2 & 62.3 \\
               &            &            & 82.8 & 60.7 \\
  \bottomrule
  \end{tabular}
\end{table}

\subsection{Ablation Studies}

We validate our core loss designs on the complex \textit{Room} scene in \cref{tab:ablation}. Omitting the scale-balancing re-weighting drops mAcc by 6.9\% and mIoU by 3.7\%, as gradients become dominated by massive entities, hindering the feature learning of fine-grained objects. Removing the auxiliary $\mathcal{L}_{pos}$ decreases mIoU by 1.3\%; without this $L_2$ penalty, cosine similarity alone yields loose intra-cluster cohesion and ambiguous boundaries. Finally, discarding feature normalization ($\mathcal{L}_{3d}$, $\mathcal{L}_{2d}$) further degrades mIoU by 1.6\%. Without length regularization, the model exploits view-dependent shortcuts, causing feature decay during alpha-blending and resulting in inconsistent 3D semantic fields.

\section{Conclusion}
In this work, we introduced LEGO, a novel framework for multi-granular 3D scene decomposition and understanding. By shifting to intrinsic structural levels and utilizing a decoupled feature space, LEGO effectively prevents cross-level semantic entanglement and ensures consistent open-vocabulary supervision. Extensive experiments demonstrate state-of-the-art performance on both promptable and open-vocabulary benchmarks, uniquely excelling at extreme fine-grained part localization. Furthermore, by structuring these hierarchies into 3D scene graphs, LEGO enables LLM-guided Chain-of-Retrieval, successfully bridging the gap between flat 3D semantic fields and complex spatial reasoning.

\section*{Acknowledgements}
This research was jointly supported by the NSFC Ph.D. Project (No. 424B2012) and NSFC General Project (No. 42571521).

%
%
\bibliographystyle{splncs04}
\bibliography{main}

\clearpage
\appendix
\section*{Supplementary Material}

The supplementary material is structured into three main sections. \Cref{sec:A} provides the algorithmic details of our methodology, including the formalization of the level-wise mask indicator, hierarchical feature learning, and the LLM-guided Chain-of-Retrieval. \Cref{sec:B} delves into the implementation details regarding the architecture, training efficiency, and evaluation protocols. \Cref{sec:C} presents additional qualitative and quantitative experimental results and analyses across diverse datasets.

\section{Algorithmic Details}
\label{sec:A}

\subsection{Formalization of Level-wise Mask Indicator}
\label{supp_sec:mask_indicator}

In the main text (Sec. 3.1), we define the level-wise mask dense indicator $\mathbb{I}_k(i, j)$ to determine whether a pixel pair $(i, j)$ belongs to the same semantic entity at a target structural level $k$. The core of this indicator relies on constructing the active mask set $\mathcal{M}_k(i)$ for each pixel $i$, translating the hierarchical axioms into formal set-theoretic operations.

Let $\mathcal{M}(i)$ denote the complete set of valid SAM masks covering pixel $i$ in the current view. Each mask $m \in \mathcal{M}(i)$ is associated with a pre-assigned structural level $l(m) \in \{1, 2, \dots, L\}$. For a given target supervision level $k$, we partition $\mathcal{M}(i)$ into two disjoint subsets: the coarser-or-equal subset $\mathcal{M}^{\leq k}(i) = \{m \in \mathcal{M}(i) \mid l(m) \leq k\}$ and the finer subset $\mathcal{M}^{> k}(i) = \{m \in \mathcal{M}(i) \mid l(m) > k\}$.

\textbf{Recursive Inclusion.} 
If a pixel is covered by a mask at a finer granularity (i.e., $l(m) > k$), it implies the pixel belongs to a sub-part of the underlying entity at level $k$. According to the recursive inclusion axiom, sharing a fine-grained identity strictly necessitates identity at coarser parent levels. Therefore, all masks in the finer subset are kept active:
\begin{equation}
    \mathcal{M}_{\text{inc}}^k(i) = \mathcal{M}^{> k}(i).
\end{equation}

\textbf{Structural Inheritance.}
Since re-graded masks are inherently sparse at any single level, a pixel might not be covered by any mask at level $k$. In this case, the pixel inherits the semantic label from its nearest valid coarser level. Mathematically, the inherited active mask set $\mathcal{M}_{\text{inh}}^k(i)$ retains masks belonging to this maximum available coarser level $l^*(i, k)$:
\begin{equation}
\begin{gathered}
\mathcal{M}_{\text{inh}}^k(i) = 
\begin{cases}
\{ m \in \mathcal{M}^{\leq k}(i) \mid l(m) = l^*(i, k) \} & \text{if } \mathcal{M}^{\leq k}(i) \neq \emptyset \\
\emptyset & \text{otherwise,}
\end{cases} \\
\text{where} \quad l^*(i, k) = \max \{ l(m) \mid m \in \mathcal{M}^{\leq k}(i) \}.
\end{gathered}
\end{equation}

\textbf{Indicator Construction.} 
By unifying the inherited and inclusive masks, we obtain the complete active mask set for pixel $i$ at level $k$:
\begin{equation}
    \mathcal{M}_k(i) = \mathcal{M}_{\text{inh}}^k(i) \cup \mathcal{M}_{\text{inc}}^k(i).
\end{equation}
Finally, the dense indicator for any sampled pixel pair $(i, j)$ is computed via the intersection of their active sets, which derives Eq. 2 in the main text:
\begin{equation}
\mathbb{I}_k(i, j) =
\begin{cases}
1 & \text{if } \mathcal{M}_k(i) \cap \mathcal{M}_k(j) \neq \emptyset \\
0 & \text{otherwise}.
\end{cases}
\end{equation}

\subsection{Details in Level-wise Feature Learning}
\label{supp_sec:feature_learning}This section provides details of the loss function and training strategy for the hierarchical feature learning omitted from the main text (Sec. 3.2) for brevity.

\textbf{Contrastive Loss.}
In the main text, the contrastive loss utilizes a scale-balancing weight $w^{i,j}_k \propto (\mathcal{A}^i_k \cdot \mathcal{A}^j_k)^{-1}$ to prevent massive entities from dominating the gradients. In practice, we normalize these weights independently within each level $k$:

\begin{equation}
\begin{gathered}
w^{i,j}_k = \frac{\tilde{w}^{i,j}_k}{\sum_{(u,v) \in \mathcal{B}} \tilde{w}^{u,v}_k}, \\
\text{where} \quad \tilde{w}^{i,j}_k = \min \left( \frac{1}{\mathcal{A}^i_k \cdot \mathcal{A}^j_k}, \tau_{\max} \right).
\end{gathered}
\end{equation}

Here, $\mathcal{B}$ represents the final subset of pixel pairs used for loss computation in the current training viewpoint (detailed in the \textbf{Pixel Pairs Sampling} section below), and $\tau_{\max}$ is a threshold preventing extremely small masks from causing exploding gradients. This intra-level normalization yields two critical optimization benefits: (1) It balances the feature optimization states between large and small objects within each level; (2) It ensures the total loss weight for every level strictly sums to 1, guaranteeing that all structural levels converge at a uniform, balanced pace despite variations in mask granularity.

Moreover, inspired by~\cite{hamilton2022unsupervised_sementic_segmentation}, we utilize a zero-clamp on the cosine similarity of negative pairs. This avoids the co-linearity and instability caused by forcing multiple distinct entities towards exactly opposite directions of a shared reference pixel in a shared feature subspace.

Overall, the complete contrastive loss is formally defined as:
\begin{equation}
\begin{gathered}
    \mathcal{L}_{contrast} = w^{i,j}_k \left( 1 - 2 \cdot \mathbb{I}_k(i, j) \right) S(\mathbf{F}_{v,k}^i, \mathbf{F}_{v,k}^j),\\
    \text{where} \quad S(\mathbf{x}, \mathbf{y}) = \begin{cases} \cos(\mathbf{x}, \mathbf{y}) & \text{if } \mathbb{I}_k(i, j) = 1, \\ \max \left( 0, \cos(\mathbf{x}, \mathbf{y}) \right) & \text{if } \mathbb{I}_k(i, j) = 0. \end{cases}
\end{gathered}
\end{equation}

\textbf{Pixel Pairs Sampling.}
While the area-based loss reweighting balances gradient magnitudes across scales, extremely fine-grained sub-parts in complex scenes may still suffer from optimization instability due to sparse spatial sampling (i.e., receiving zero or very few pixel hits in a given iteration). To address this, alongside uniform sampling, we dynamically sample an additional set of pixels weighted by their underlying semantic scales. Specifically, the sampling probability $p(i)$ for a pixel $i$ is formulated by normalizing the sum of its inverse active mask areas across all $L$ levels:
\begin{equation}
    p(i) = \frac{\sum_{k=1}^L (\mathcal{A}^i_k)^{-1}}{\sum_{j \in \mathcal{V}} \sum_{k=1}^L (\mathcal{A}^j_k)^{-1}},
\end{equation}

where $\mathcal{V}$ denotes the set of all valid pixels in the current view.

We then form $N_p^2 / 2$ unique pixel pairs based on the $N_p$ sampled pixels. To further balance the optimization of positive and negative pairs, inspired by~\cite{cen2025SAGA}, we utilize our dense indicator $\mathbb{I}_k(i, j)$ to partition the pairs into three mutually exclusive sets based on their hierarchical relationships:

\begin{itemize}
\item \textit{Inconsistent Set} ($Q_{in}$): Pairs sharing identity at coarse levels but splitting at finer levels, formulated as $Q_{in} = \{ (i, j) \mid \exists k_1, k_2, \mathbb{I}_{k_1}(i, j) \neq \mathbb{I}_{k_2}(i, j) \}$.
\item \textit{Consistent Positive Set} ($Q_{pos}$): Pairs maintaining identical semantics across all levels, formulated as $Q_{pos} = \{ (i, j) \mid \forall k, \mathbb{I}_k(i, j) = 1 \}$.
\item \textit{Consistent Negative Set} ($Q_{neg}$): Pairs that never share an identity at any level, formulated as $Q_{neg} = \{ (i, j) \mid \forall k, \mathbb{I}_k(i, j) = 0 \}$.
\end{itemize}

In each training iteration, we include all pairs from $Q_{in}$ and randomly sample $|Q_{in}|/2$ pairs from both $Q_{pos}$ and $Q_{neg}$ to form the final computation batch $\mathcal{B}$ for computing $\mathcal{L}_{contrast}$ and $\mathcal{L}_{pos}$.

\subsection{Details in LLM-Guided Chain-of-Retrieval}
\label{supp_sec:cor}

In the main text (Sec. 3.3), we organize the hierarchical 3D segments into a language scene graph with Hierarchical Edges ($\mathcal{E}_{hier}$) and Spatial Adjacency Edges ($\mathcal{E}_{adj}$). This section details the subsequent LLM-guided Chain-of-Retrieval (CoR) algorithm.

\textbf{Graph Edge Refinement.} Since parent and child nodes naturally intersect in 3D space, they inherently trigger both $\mathcal{E}_{hier}$ and $\mathcal{E}_{adj}$. To prevent redundant pathways and maintain topological clarity, we remove any $\mathcal{E}_{adj}$ between nodes already connected by $\mathcal{E}_{hier}$.

\textbf{LLM Query Parsing.}
When faced with a complex human instruction, we prompt an LLM agent to parse the raw text into an ordered sequence of target objects $\mathcal{O} = (o_1, o_2, \dots, o_T)$ and their corresponding relations $\mathcal{R} = (r_1, r_2, \dots, r_{T-1})$, where $r_t \in \{\text{hier}, \text{adj}\}$. For instance, the query \textit{``Find the handle of the pitcher beside the rolling pin''} is parsed into a coarse-to-fine sequence: $\mathcal{O} = [\text{rolling pin}, \text{pitcher}, \text{handle}]$ and relations $\mathcal{R} = [\text{adj}, \text{hier}]$.

\textbf{Relation-Constrained Beam Search.}
Guided by this parsed sequence, we execute a sequential Beam Search over the Scene Graph $\mathcal{G}$. Let $\Psi_t = (v_1, v_2, \dots, v_t)$ denote a retrieval path at step $t$. The cumulative semantic score of this path is defined as the sum of cosine similarities along the chain:
\begin{equation}
    S(\Psi_t) = \sum_{k=1}^t \cos(\mathbf{E}_{v_k}, \mathbf{E}_{o_k}),
\end{equation}

where $\mathbf{E}_{v_k}$ is the view-aware CLIP feature of node $v_k$, and $\mathbf{E}_{o_k}$ is the text embedding of the parsed sub-query object $o_k$.

The retrieval initializes at $t=1$ by selecting the top-$K$ nodes  \textbf{$v \in \mathcal{V}$} (where $K$ is the beam width) in the graph that maximize $\cos(\mathbf{E}_{v}, \mathbf{E}_{o_1})$ as entry points. For each subsequent step $t \in \{2, \dots, T\}$, the current paths are iteratively expanded to the neighborhood of the previous node. Crucially, the valid candidate neighborhood $\mathcal{N}_{r_{t-1}}(v_{t-1})$ is strictly constrained by the parsed relation $r_{t-1}$: expansion is \textbf{only permitted} through $\mathcal{E}_{hier}$ if $r_{t-1} = \text{hier}$, or through $\mathcal{E}_{adj}$ if $r_{t-1} = \text{adj}$.

After computing the updated cumulative scores $S(\Psi_t)$ for all candidate branches, we prune the search tree by retaining only the top-$K$ paths to maintain computational efficiency. Once the entire sub-query chain is traversed, the terminal node $v_T$ of the path with the highest cumulative score $S(\Psi_T)$ is successfully retrieved as the definitive target. By explicitly incorporating graph topology, this chaining paradigm fundamentally overcomes the ambiguity of flat semantic matching, enabling precise, context-aware visual grounding.

\section{Implementation Details}
\label{sec:B}

\textbf{Architecture and Training Efficiency.} 
Following previous approaches, we employ SAM (ViT-H) \cite{kirillov2023sam} to construct the 2D mask pool and build our Gaussian Splatting framework based on the implementation from gsplat \cite{ye2025gsplat}. For each scene, we first optimize the standard RGB Gaussian field for $30k$ iterations. Then, we freeze all geometric and appearance attributes to exclusively optimize the identity features for an additional $10k$ iterations. Across all experiments, we configure the identity feature with a maximum depth of $L=8$ levels, allocating $d=8$ dimensions per level. This configuration stems from our empirical observation that real-world entities rarely decompose beyond 8 granular levels, rendering this capacity fully sufficient for complex scenes.

To accelerate the feature learning stage, we randomly sample 4 levels and 1,500 pixels per image in each iteration, yielding $1,500^2 / 2$ pixel pairs. Crucially, because our feature space is structurally decoupled, the renderer only computes and back-propagates features for the sampled levels rather than the entire $L$-level hierarchy. This sparse rendering strategy successfully reduces computational overhead. Consequently, the complete two-stage Gaussian field training requires $\sim$20–60 minutes per scene on a single RTX 4090 GPU. Furthermore, by extracting CLIP features exclusively on the optimal views of each decoupled object, we avoid redundant dense feature rendering. This efficient extraction strategy limits the subsequent tree-structured scene decoupling and language feature extraction to an additional $\sim$5–10 minutes.

\textbf{Evaluation Protocols.} 
To translate our 3D hierarchical representation into quantitative 2D metrics, we establish specific evaluation protocols tailored to different benchmark settings:

\begin{itemize}
    \item \textit{Promptable Segmentation:} For the NVOS dataset, we unproject the provided 2D scribbles from the reference view into the decoupled 3D Gaussian field using camera parameters. We identify the target 3D cluster by maximizing the feature similarity margin between the positive and negative projected points. The selected cluster is then rendered to the target views to compute mIoU and mAcc metrics against the ground truth. We apply a similar unprojection strategy for the SPIn-NeRF dataset to extend reference-view masks to all evaluation views.
    
    \item \textit{Open-Vocabulary Understanding:} Since the LERF-OVS and Mip-NeRF 360 datasets provide only single-word or phrase queries, we perform evaluation using solely the grounded CLIP features, without invoking the constructed scene graph or the LLM-based Chain-of-Retrieval (CoR). Specifically, following the evaluation protocols of prior baselines, we compute the cosine similarity between the CLIP embeddings of the text query and the 3D clusters within the top three levels of our hierarchy. We first identify the specific level containing the cluster with the highest similarity score in the evaluation view, denoted as $S_{\max}$. Recognizing that multiple instances of a queried object may exist within a single view, we robustly retrieve all clusters \textit{within that selected level} exhibiting a similarity score $S \geq 0.9 \cdot S_{\max}$. For the segmentation task, these matched clusters are projected onto the 2D plane for metric evaluation. For the localization task, we render the similarity map of the 3D clusters onto the 2D plane and extract the pixel position with the maximum value as the predicted target location.
\end{itemize}
\section{Additional Results}
\label{sec:C}

\begin{table}[ht]
  \caption{Open-vocabulary 3D understanding results on 3D-OVS dataset.}
  \label{tab:result_3dovs}
  \centering
  \setlength{\tabcolsep}{4pt}
  \begin{tabular}{l|cccccc}
    \toprule
    Method & \textit{bed} & \textit{bench} & \textit{room} & \textit{sofa} & \textit{lawn} & \textit{\textbf{Overall}} \\ 
    \midrule  
    3D-OVS \cite{liu20233dovs} & 89.5 & 89.3 & 92.8 & 74.0 & 88.2 & 86.8 \\
    LEGaussians \cite{shi2024legaussian} & 84.9 & 91.1 & 86.0 & 87.8 & 92.5 & 88.5 \\
    GOI \cite{qu2024goi} & 89.4 & 92.8 & 91.3 & 85.6 & 94.1 & 90.6 \\ 
    LangSplat \cite{qin2024langsplat} & 92.5 & 94.2 & 94.1 & 90.0 & 96.1 & 93.4 \\
    N2F2 \cite{bhalgat2024n2f2} & 93.8 & 92.6 & 93.5 & 92.1 & 96.3 & 93.9 \\
    LangSplatV2 \cite{qin2024langsplat} & 93.0 & 94.9 & 96.1 & 92.3 & 96.6 & 94.6 \\
    Occam’s LGS \cite{cheng2024occamlgs} & 96.8 & \textbf{95.8} & 96.5 & 88.8 & \underline{97.0} & 95.0 \\
    LaGa \cite{cen2025laga} & 96.8 & 92.8 & \textbf{97.0} & 93.0 & 96.9 & 95.3 \\
    SAGA \cite{cen2025SAGA} & \underline{97.4} & \underline{95.4} & \underline{96.8} & \underline{93.5} & 96.6 & \underline{96.0} \\
    \midrule
    \textbf{LEGO(Ours)} & \textbf{97.5} & \textbf{95.8} & \underline{96.8} & \textbf{95.0} & \textbf{97.3} & \textbf{96.5} \\
  \bottomrule
  \end{tabular}
\end{table}

\subsection{Quantitative Results}

In addition to the benchmarks reported in the main text, we further evaluate LEGO on the 3D-OVS dataset \cite{liu20233dovs} to validate its open-vocabulary 3D understanding capabilities. As shown in \cref{tab:result_3dovs}, LEGO achieves an overall performance of 96.5\% mIoU, outperforming recent competitive baselines such as LaGa~\cite{cen2025laga} and SAGA~\cite{cen2025SAGA}. Notably, LEGO secures the highest accuracy in four out of the five evaluated scenes (\textit{bed}, \textit{bench}, \textit{sofa}, and \textit{lawn}). These quantitative improvements demonstrate that LEGO not only produces robust semantic matching but also yields more accurate boundaries across diverse open-world scenarios.

\begin{figure}[ht!]
  \centering
  \includegraphics[width=\linewidth]{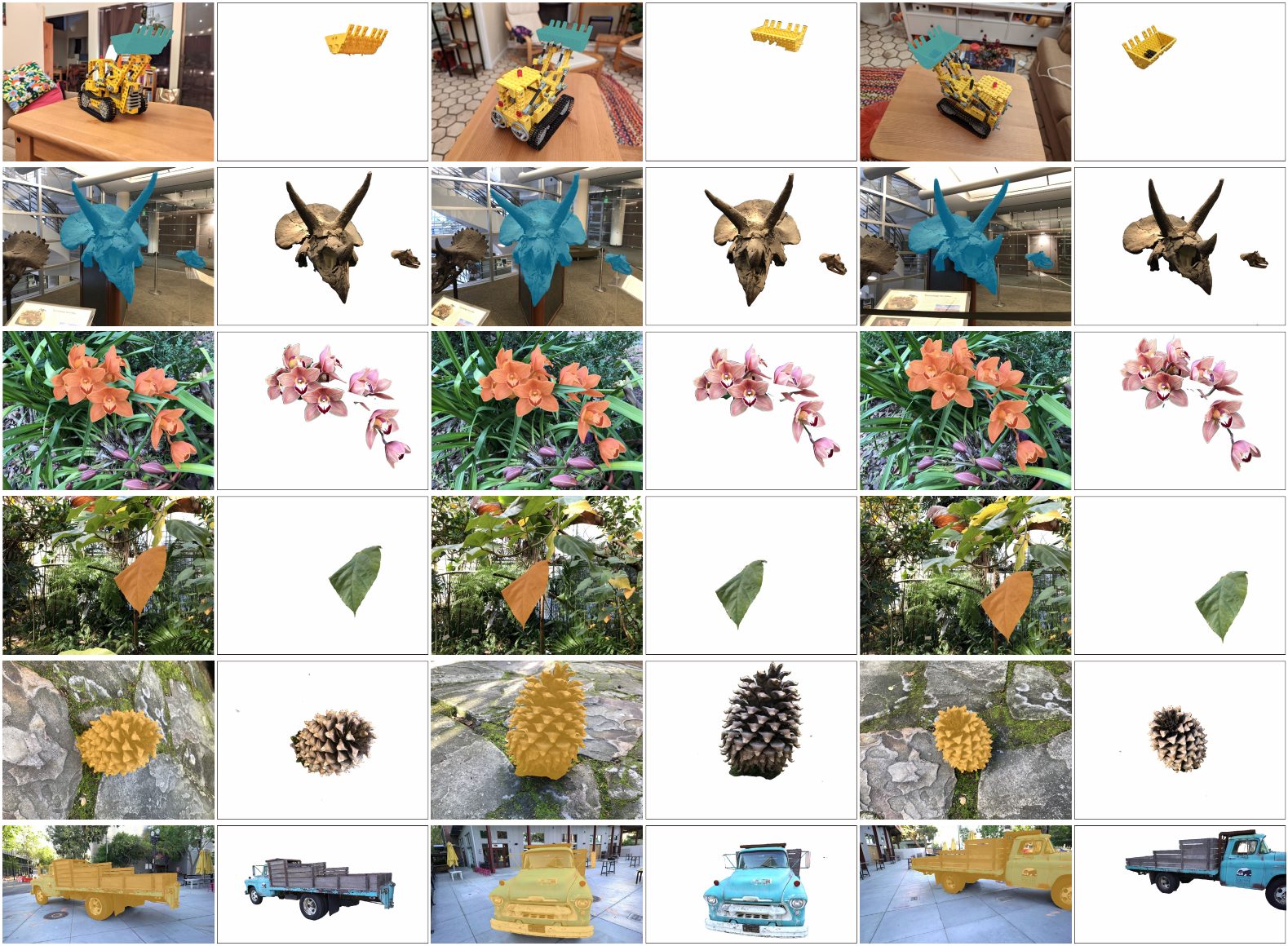}
  \caption{\textbf{Additional promptable segmentation results in different scenes.} LEGO consistently produces sharp and accurate object boundaries across various complex topologies.}
  \label{fig:ps_vis_main}
\end{figure}

\begin{figure}[ht!]
  \centering
  \includegraphics[width=\linewidth]{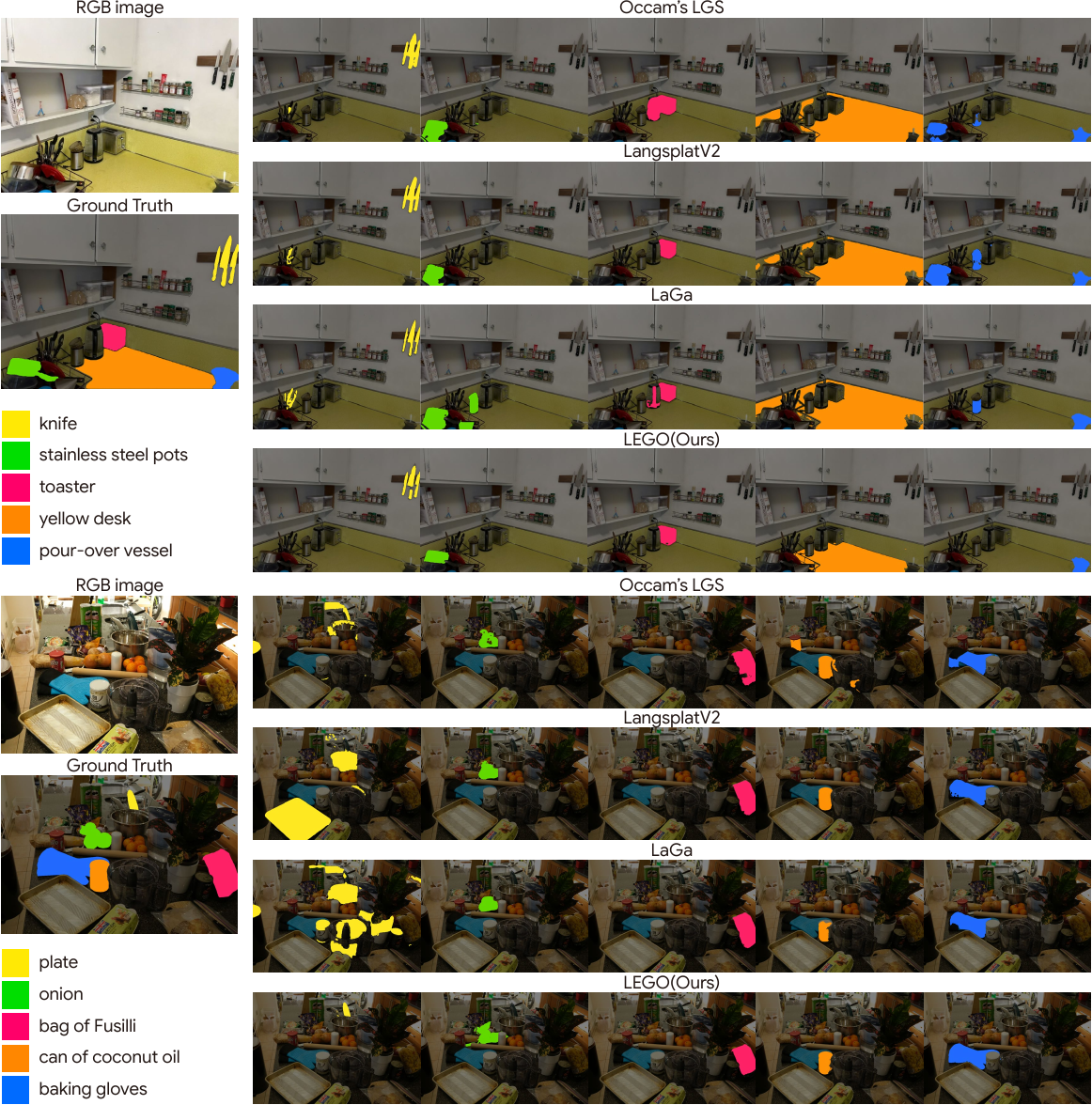}
  \caption{\textbf{Additional open-vocabulary segmentation results in \textit{Kitchen} and \textit{Counter} scene.} LEGO demonstrates superior robustness in highly cluttered environments. It accurately isolates the \textit{``yellow desk''} from a visually similar background wall, and successfully segments \textit{``plate''} and all five \textit{``onions''}.}
  \label{fig:ovs_vis_main_2}
\end{figure}

\begin{figure}[ht!]
  \centering
  \includegraphics[width=\linewidth]{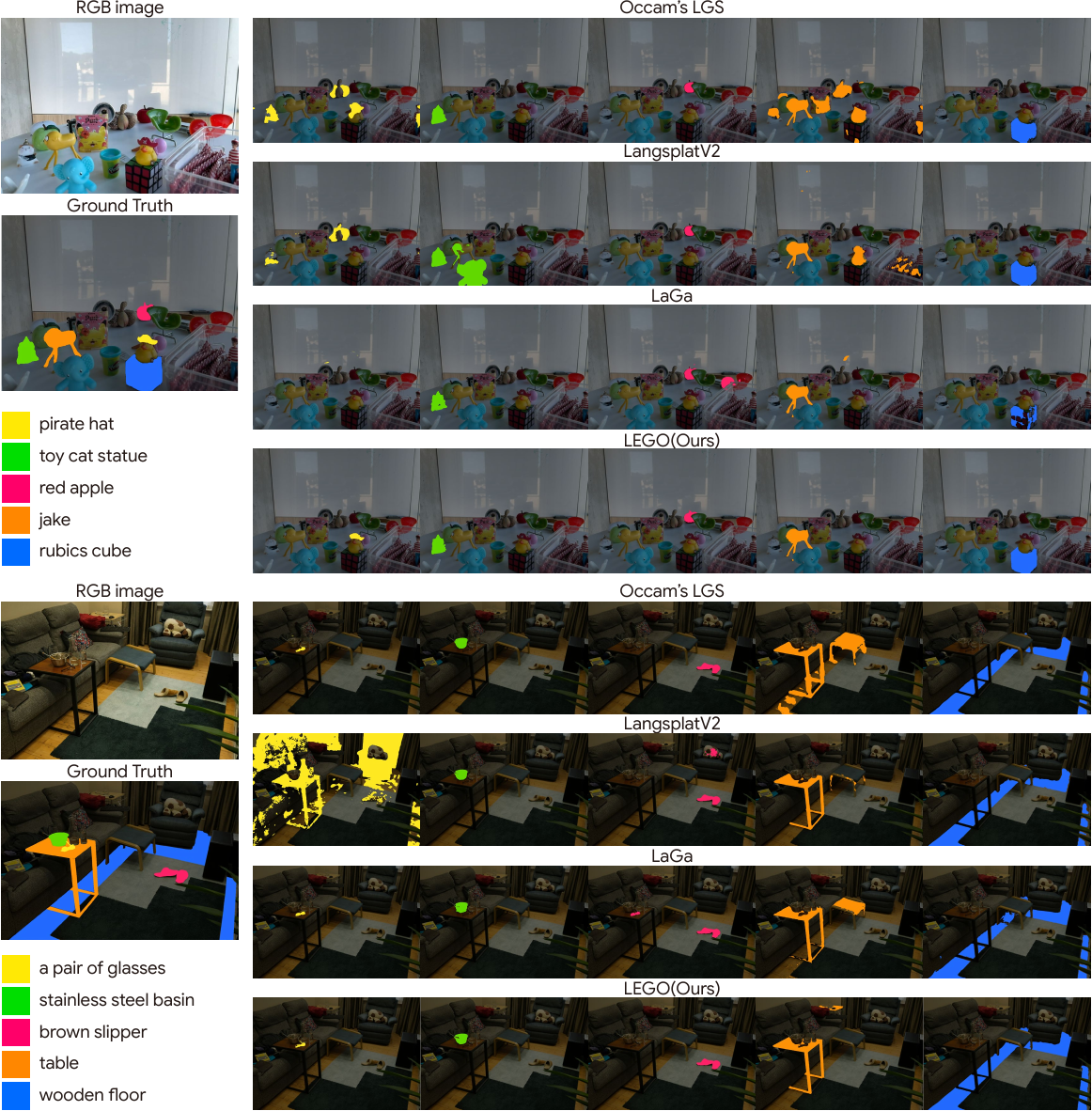}
  \caption{\textbf{Additional open-vocabulary segmentation results in \textit{Figurines} and \textit{Room} scene.} LEGO cleanly segments extremely fine-grained component \textit{``pirate hat''}, and yields the most complete regions for \textit{``wooden floor''}. Remarkably, it even retrieves unannotated \textit{``table''} behind the sofa. }
  \label{fig:ovs_vis_main_3}
\end{figure}

\begin{figure}[ht!]
  \centering
  \includegraphics[width=0.75\linewidth]{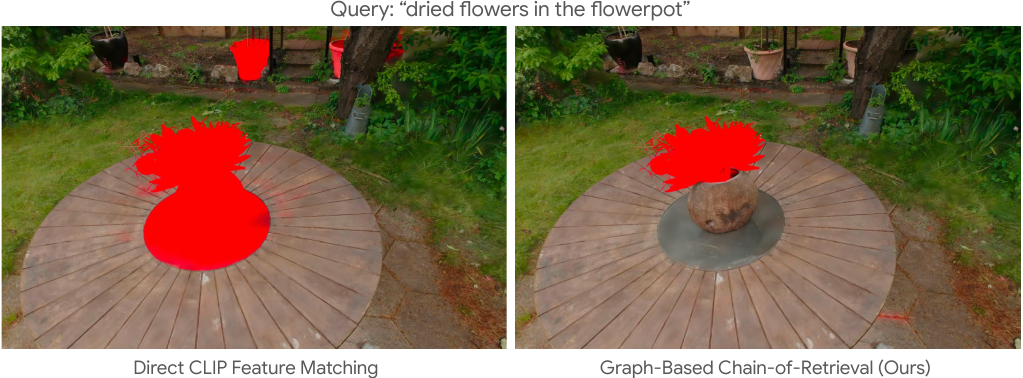}
  \caption{\textbf{Effectiveness of Graph-Based Chain-of-Retrieval (CoR).} When given a compositional query \textit{``dried flowers in the flowerpot''}, direct CLIP feature matching \textbf{(left)} suffers from the ``bag-of-words'' effect, erroneously activating the pot on the table as well as irrelevant background pots alongside the target. In contrast, our CoR paradigm \textbf{(right)} leverages the hierarchical scene graph to precisely isolate the intended object, resolving the semantic ambiguity of entangled text embeddings.}
  \label{fig:ablation_COR}
\end{figure}

\begin{figure}[ht!]
  \centering
  \includegraphics[width=\linewidth]{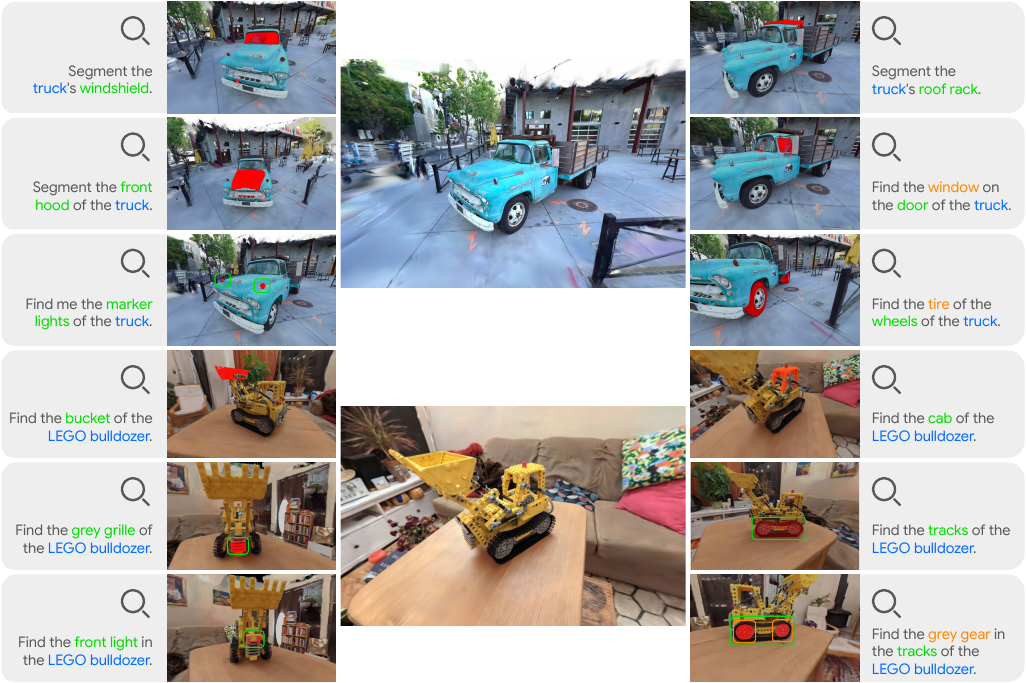}
  \caption{\textbf{Additional LLM-based chain of retrieval (CoR) results.} LEGO accurately grounds corresponding functional sub-parts (\eg, \textit{lights}, \textit{wheels}, \textit{hood}) via language, demonstrating robust semantic understanding across a massive real-world truck and a miniature LEGO bulldozer.}
  \label{fig:ovs_vis_llm_query_2}
\end{figure}

\begin{figure}[ht!]
  \centering
  \includegraphics[width=\linewidth]{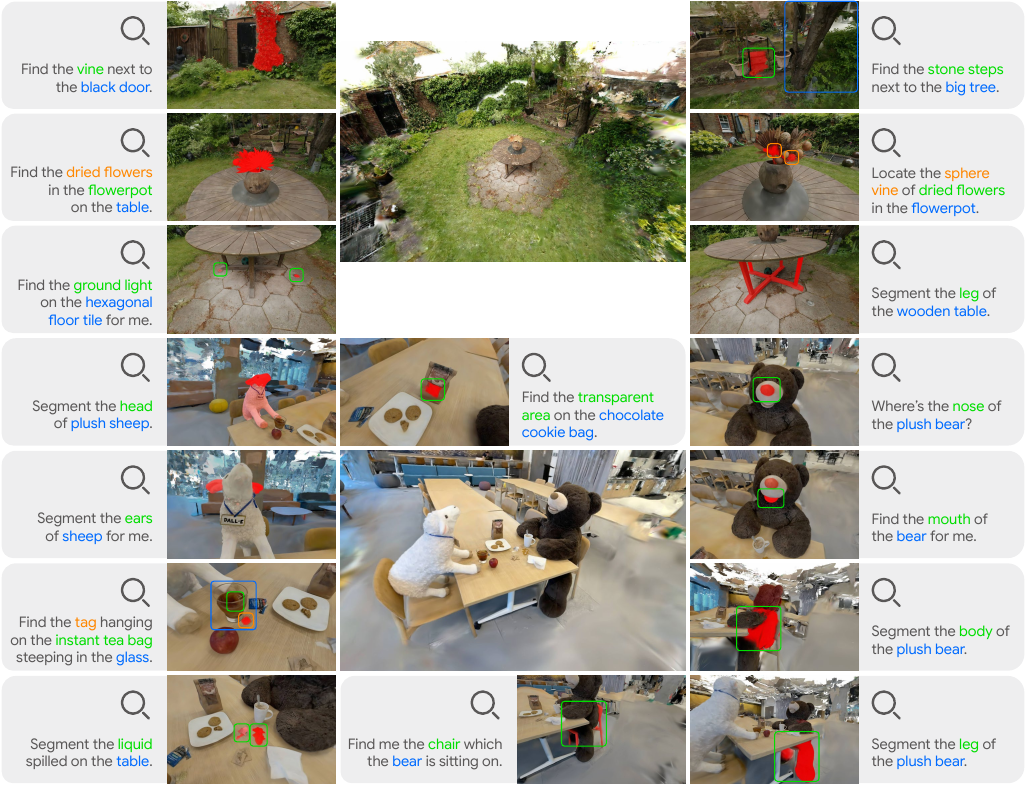}
  \caption{\textbf{Additional LLM-based chain of retrieval (CoR) results.} Our method successfully handles extreme scale variations and complex queries. It localizes tiny objects (\eg, \textit{ground lights}), resolves deep nested semantics (\eg, \textit{flowerpot} $\to$ \textit{dried flowers} $\to$ \textit{sphere vine}), and distinguishes specific sub-parts of plush bear and sheep.}
  \label{fig:ovs_vis_llm_query_3}
\end{figure}

\begin{figure}[ht!]
  \centering
  \includegraphics[width=\linewidth]{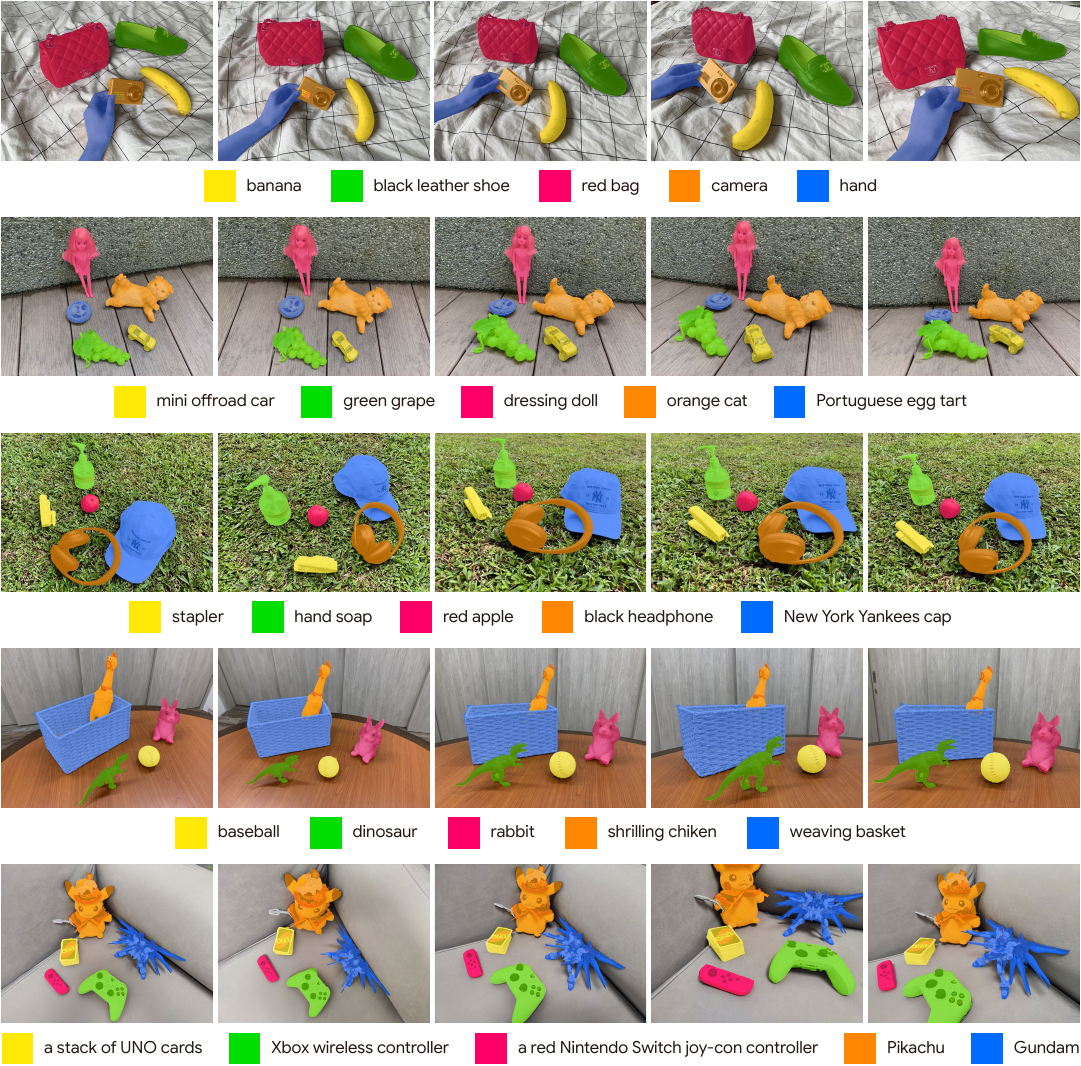}
  \caption{\textbf{Additional open-vocabulary segmentation results in \textit{Bed}, \textit{Bench}, \textit{Lawn}, \textit{Room} and \textit{Sofa} scenes of the 3D-OVS dataset.} }
  \label{fig:ovs_vis_main_3dovs}
\end{figure}

\subsection{Qualitative Results}

In this section, we provide extended qualitative visualizations. Specifically, \cref{fig:ps_vis_main} presents visual results for promptable 3D segmentation; \cref{fig:ovs_vis_main_2} and \cref{fig:ovs_vis_main_3} showcase additional comparisons for open-vocabulary segmentation across diverse scenes; \cref{fig:ovs_vis_llm_query_2} and \cref{fig:ovs_vis_llm_query_3} demonstrate our graph-based Chain-of-Retrieval capabilities; \cref{fig:ovs_vis_main_3dovs} visualizes the results on the 3D-OVS dataset.

\textbf{Promptable Segmentation.}
As illustrated in \cref{fig:ps_vis_main}, we present additional qualitative results for the promptable segmentation task across a diverse array of indoor and outdoor scenes. LEGO consistently produces highly accurate 3D segmentations with exceptionally sharp and exquisite boundaries, even for objects with complex topologies (\eg, the delicate edge of the \textit{Orchids} and \textit{Horns}, or the slim rearview mirror structure of the \textit{Truck}). This high-fidelity boundary preservation is fundamentally attributed to the purity of our decoupled feature field. By explicitly isolating multi-granular features into independent hierarchical levels and strictly enforcing spatial constraints via dense indicators, LEGO effectively prevents cross-level feature smoothing and spatial bleeding.

\textbf{Open-Vocabulary Segmentation.}
As demonstrated in \cref{fig:ovs_vis_main_2} and \cref{fig:ovs_vis_main_3}, LEGO mitigates the feature ambiguity and bleeding that plague baseline methods. In the \textit{Kitchen} and \textit{Room} scenes, LEGO accurately isolates the \textit{``yellow desk''} from the visually similar background wall, cleanly extracts \textit{``stainless steel pots''} from heavy clutter, and segments the most intact \textit{``wooden floor''}.

Furthermore, our method excels at localizing fine-grained objects. In the \textit{Figurines} scene, LEGO uniquely succeeds in segmenting the fine-grained component \textit{``pirate hat''} atop the rubber duck. In the \textit{Counter} scene, it not only localizes the occluded \textit{``plate''} but also precisely identifies all five \textit{``onions''}, whereas alternative approaches miss one or more of them due to under-segmentation.

Remarkably, in the \textit{Room} scene, when queried for \textit{``table''}, LEGO not only segments the primary target but also successfully retrieves an additional, unannotated table hidden behind the sofa. These results consistently validate that our optimal-view feature extraction and level-wise 3D architecture enable robust, precise, and true open-world scene understanding.

Moreover, \cref{fig:ovs_vis_main_3dovs} visualizes LEGO's performance on the 3D-OVS dataset. These results highlight LEGO's exceptional capability in segmenting highly diverse and topologically complex entities. For instance, LEGO flawlessly isolates intricate structures with sharp geometric details, such as the spikes of the \textit{``Gundam''} and the hollowed structure of the \textit{``weaving basket''}. Across all varied viewpoints, the segmented masks maintain exquisite multi-view consistency and strict boundary adherence.

\textbf{Graph-Based Chain-of-Retrieval.}
As illustrated in \cref{fig:ovs_vis_llm_query_2} and \cref{fig:ovs_vis_llm_query_3}, we present comprehensive qualitative results of our LLM-guided Chain-of-Retrieval (CoR) across scenes with diverse scales and types. In \cref{fig:ovs_vis_llm_query_2}, despite the massive scale and domain gap between a real outdoor truck and a miniature LEGO bulldozer on the table, our method demonstrates remarkable cross-domain semantic correspondence, accurately retrieving functional sub-parts (\eg, \textit{marker lights}, \textit{wheels}, \textit{front hood}) based solely on language queries.

\cref{fig:ovs_vis_llm_query_3} further validates our robustness against drastic scale variations and complex nested semantics. LEGO seamlessly localizes tiny entities like \textit{ground light} in a sprawling garden, distinguishes specific fine-grained parts of plush toys (\eg, \textit{nose}, \textit{mouth}, \textit{legs}), and accurately parses multi-layered semantic inclusions (\eg, retrieving the \textit{sphere vine} within the \textit{dried flowers} inside a \textit{flowerpot}).

Crucially, these results highlight the fundamental advantage of our CoR paradigm over flat CLIP-based matching. Standard vision-language foundation models notoriously suffer from the ``bag-of-words'' effect. As compared in \cref{fig:ablation_COR}, when presented with a compositional prompt like \textit{``the dried flowers in the flowerpot''}, the entangled text embedding yields high similarities for both the flowers and the pot, leading to ambiguous localization. By employing an LLM to decompose long instructions into atomic queries and executing them sequentially over our hierarchical scene graph, LEGO elegantly bypasses this semantic entanglement. Furthermore, this explicit node-edge topology natively paves the way for future extensions into even more sophisticated spatial reasoning tasks, such as distance-based or orientation-aware grounding.

\end{document}